\documentclass[10pt,twocolumn,letterpaper]{article}
\usepackage[letterpaper,top=1in,left=0.8125in,textwidth=6.875in,textheight=8.875in,columnsep=0.3125in]{geometry}
\usepackage[T1]{fontenc}
\usepackage{times}

\usepackage{graphicx}
\usepackage{amsmath}
\usepackage{amssymb}
\usepackage{booktabs}
\usepackage{multirow}
\usepackage{xcolor}
\usepackage{pifont}
\usepackage[font=small,labelfont=bf,labelsep=colon]{caption}
\usepackage[numbers,sort&compress]{natbib}
\usepackage{titlesec}
\titleformat{\section}{\large\bfseries}{\thesection}{0.6em}{}
\titleformat{\subsection}{\normalsize\bfseries}{\thesubsection}{0.6em}{}
\titlespacing*{\section}{0pt}{1.8ex plus .5ex minus .2ex}{0.9ex plus .2ex}
\titlespacing*{\subsection}{0pt}{1.4ex plus .4ex minus .2ex}{0.7ex plus .2ex}
\newcommand{\cmark}{\ding{51}}
\newcommand{\xmark}{\ding{55}}
\newcommand{\ns}{$^{\dagger}$}

\definecolor{linkblue}{rgb}{0.21,0.49,0.74}
\usepackage[breaklinks,colorlinks,allcolors=linkblue]{hyperref}

\title{%
  \makebox[\textwidth][c]{%
    \raisebox{-0.3cm}{\includegraphics[height=1.5cm]{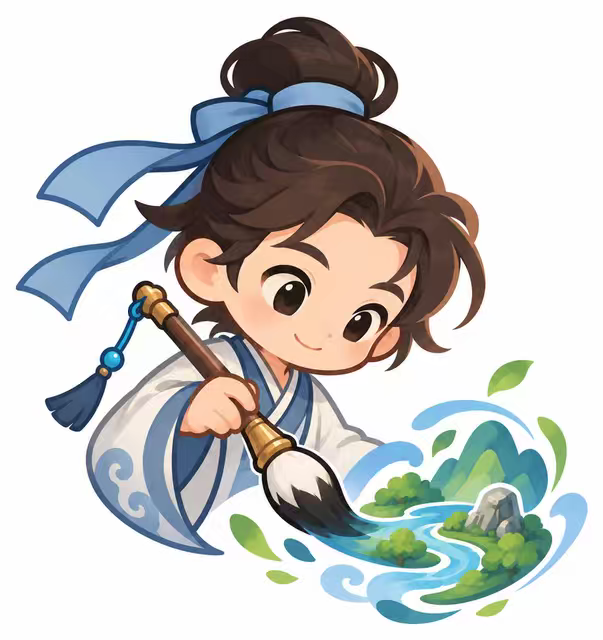}}%
    \hspace{0.1cm}%
    \mbox{\bfseries MaLiang: Painting Playable Game Worlds from Numbers}%
  }%
}

\author{Zhiyang Deng$^{1,2,\dagger}$\qquad Shushan Xu$^{1,3,\dagger}$\qquad Yeying Jin$^{1,2,*,\ddagger}$\\[0.35em]
{\normalsize $^{1}$Tencent\qquad $^{2}$National University of Singapore\qquad $^{3}$Nanyang Technological University}}
\date{}

\begin{document}
\twocolumn[{%
\renewcommand\twocolumn[1][]{#1}%
\maketitle
\vspace{-1.1em}
\begin{center}
\centering
\includegraphics[width=\linewidth]{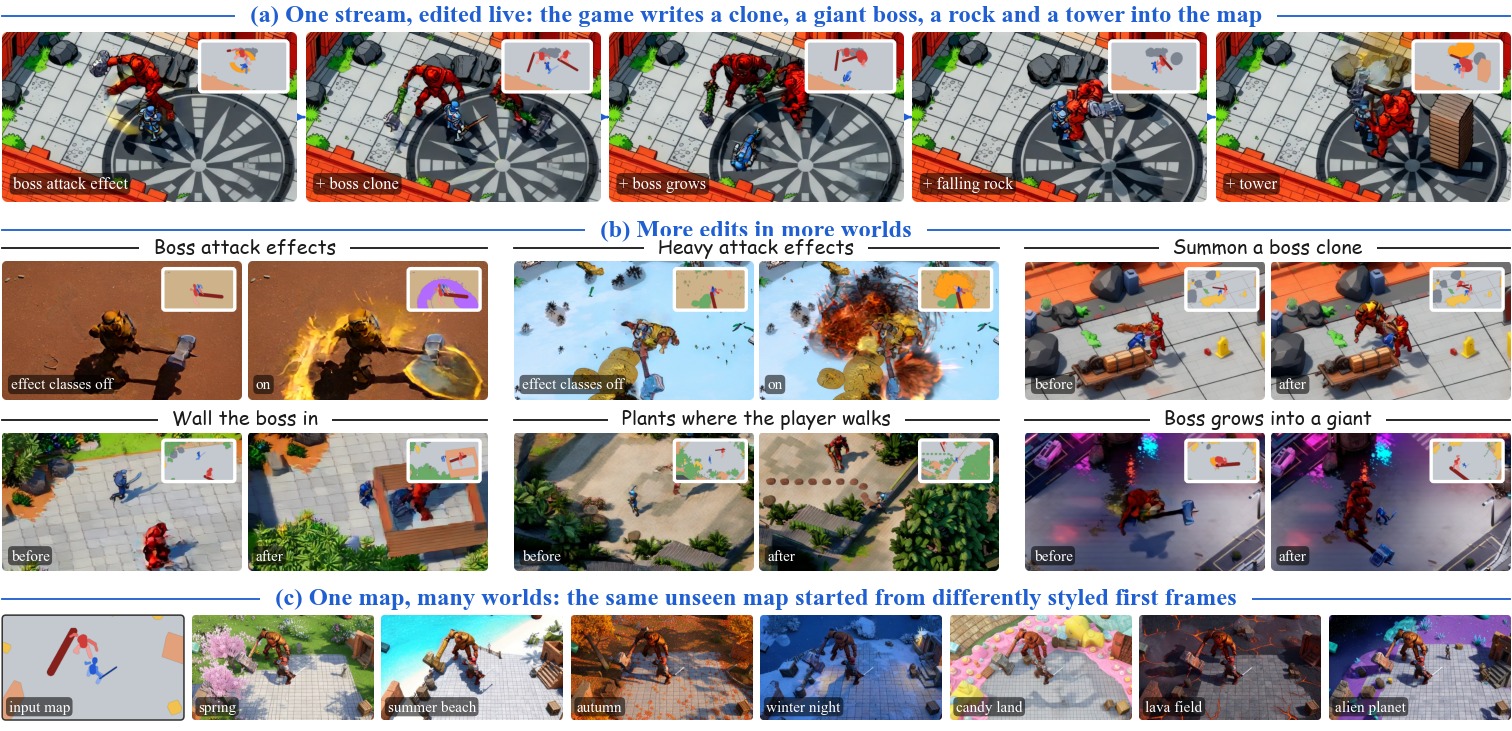}
\captionsetup{hypcap=false}
\captionof{figure}{\textbf{A game controls MaLiang by writing integers.} Every large image is generated in real time by
  our causal student from the integer map in its inset. (a)~One comic-style boss fight from our interactive demo: while
  the stream runs, game rules write a boss clone, a giant boss, a falling rock and a tower into the map, and the next
  steps follow. (b)~First two pairs: attack-effect classes switched off and on in two runs with the same seed and map; the other pairs
  are two moments of one demo stream. (c)~One unseen procedural map started from differently styled first frames (3.8\,s).
  Demo worlds start from first frames that an image model re-drew in a chosen style.}
\captionsetup{hypcap=true}
\label{fig:teaser}
\end{center}
\vspace{0.4em}
}]
\begingroup
\let\thefootnote\relax
\begin{NoHyper}
\footnotetext{$^{\dagger}$Work done during internships at Tencent, supervised by Yeying Jin.}
\footnotetext{$^{*}$Corresponding author.\quad $^{\ddagger}$Project Lead.}
\end{NoHyper}
\endgroup

\begin{abstract}
Video models now render game footage of remarkable quality, and recent systems use them as the renderer of an
executable game: a simulator keeps the world state and rules, and the video model paints the frames. Most connect the
two through a visual proxy---white-box, low-poly or colour-coded videos encoded by a video autoencoder---so every world
must first be built and drawn in an engine. MaLiang removes this detour. Its \emph{numerical proxy} is a per-pixel
integer map of the whole world, including characters, body parts and attack effects; \emph{Native Numerical
Conditioning} adds it to a video diffusion transformer through a 9.4M-parameter class-pooling encoder, with no rendered
video and no visual encoder; and the map becomes a \emph{Numerical Canvas} that game rules edit and the model paints.
With backbone and data fixed, integer conditioning matches colour-coded inputs in image quality and layout and controls
characters and effects more precisely, even against a colour encoder of identical cost. Distilled into a causal student
that streams at 22 frames per second on one GPU, MaLiang renders edits in the next step and paints unseen procedural
worlds. On held-out recordings it reaches lower FVD (154 vs.\ 391) than the strongest off-the-shelf renderer, a 33B model, at statistically tied semantic accuracy. 
Project Page: \href{https://yeyinglab.github.io/MaLiang/}{\textcolor{blue}{https://yeyinglab.github.io/MaLiang/}}
Code: \href{https://github.com/YeyingLab/MaLiang}{\textcolor{blue}{https://github.com/YeyingLab/MaLiang}}
\end{abstract}

\section{Introduction}
\label{sec:intro}

Video generation models can now render plausible game footage, and a growing line of work uses them as the renderer of
an interactive game: a conventional engine or a coding agent keeps the game state, and a video model draws it
\cite{magpie2026,matrixgame4,streamrender2026,cwm2026,pwm2026}. The interface between the two halves is almost always an
image. The state is first rendered into a visual proxy---untextured whitebox frames \cite{magpie2026}, a low-poly scene
\cite{matrixgame4}, a flat-coloured semantic video \cite{streamrender2026,solarrenderer2026}, depth plus semantics
\cite{cwm2026}, or grey blockouts with colour-coded entities \cite{anygamerenderer2026}---and the proxy is encoded by a
video autoencoder or a reference-video pathway of the generator.

This interface has three costs. \textbf{(1) Implicit categories:} the network has to recover what is where from
colours, and after downsampling a thin weapon or a short-lived effect is averaged with its surroundings; effects such as
a boss's attack have no natural proxy appearance at all---in the renderers we tested they become a generic colour or
disappear. \textbf{(2) Rigid coupling:} the proxy is a rendering style---a palette, a level of geometric detail---that
the renderer was trained on, so the game must reproduce that style exactly and every renderer needs its own proxy pass,
whereas engine ids exist in every game. \textbf{(3) Extra computation:} every step needs a rendering pass and a visual
encoder on top of the generator; a video-VAE proxy encoder alone takes 49\,ms per four-frame step on one GPU (amortised
over a clip).

We take the opposite position: \emph{the rendered semantics of a game are already discrete, so give them to the model as integers}
(Fig.~\ref{fig:teaser}).
MaLiang\footnote{Named after Ma Liang, the boy painter of a Chinese folk tale whose magic brush brings his drawings to life.} rasterises the
numeric world into one 8-bit class map per frame, its \emph{numerical proxy}, covering seven scene categories, six body
parts for each of the two characters, and four attack-effect classes. \emph{Native Numerical Conditioning} adds a
learned embedding of that map directly to the patch embedding of a pretrained video diffusion transformer
(Wan2.2-TI2V-5B \cite{wan2025}), with no proxy video and no visual encoder. A causal four-step student
\cite{causvid2025,selfforcing2025,dmd2024} then streams video in 0.25-s steps, so the map becomes a
\emph{Numerical Canvas}: a new rock or a removed tree written into it reaches the screen in the next step. The map
itself is a semantic label map, as in SPADE-style synthesis \cite{spade2019}; what is new is how it enters a latent
video transformer: as class fractions at latent resolution, grouped in time exactly like the video latent, so that each
causal step reads only its own slice of the map and edits stream step by step.

Evaluating this claim requires care, because existing renderers consume different inputs, run at different resolutions
and frame rates, and report different metrics. We therefore build one protocol for all methods: 100 five-second clips
from ten recordings never seen in training, every output resampled to a common grid, and the same appearance, semantic,
entity and temporal metrics with paired bootstrap confidence intervals. Our contributions are:
\begin{itemize}
\item \textbf{A numerical proxy with Native Numerical Conditioning}: the whole world, including body parts and attack
effects, as one integer map per frame, read by a 9.4M-parameter encoder in 0.9\,ms per step.
\item \textbf{A Numerical Canvas for playable worlds}: a causal student streams 832$\times$480 video at 22\,fps on one
GPU, so whatever game rules write into the map is painted in the next step; our browser demo is played this way.
\item \textbf{A unified comparison and a same-backbone interface ablation.} Against three structure-to-video renderers
run off the shelf on our data, MaLiang has the best FVD, LPIPS and PSNR, no significant difference in semantic accuracy,
and about 30$\times$ the frame rate of their whole-clip configurations; with the
backbone, data and training fixed, integer conditioning matches colour-image conditioning in layout control and image quality
and is significantly more precise on characters and attack effects, a gap that widens when both are distilled into real-time students with the same distillation budget.
\end{itemize}
Code, the deploy model and the test set will be released.

\begin{figure*}[t]
  \centering
  \includegraphics[width=\linewidth]{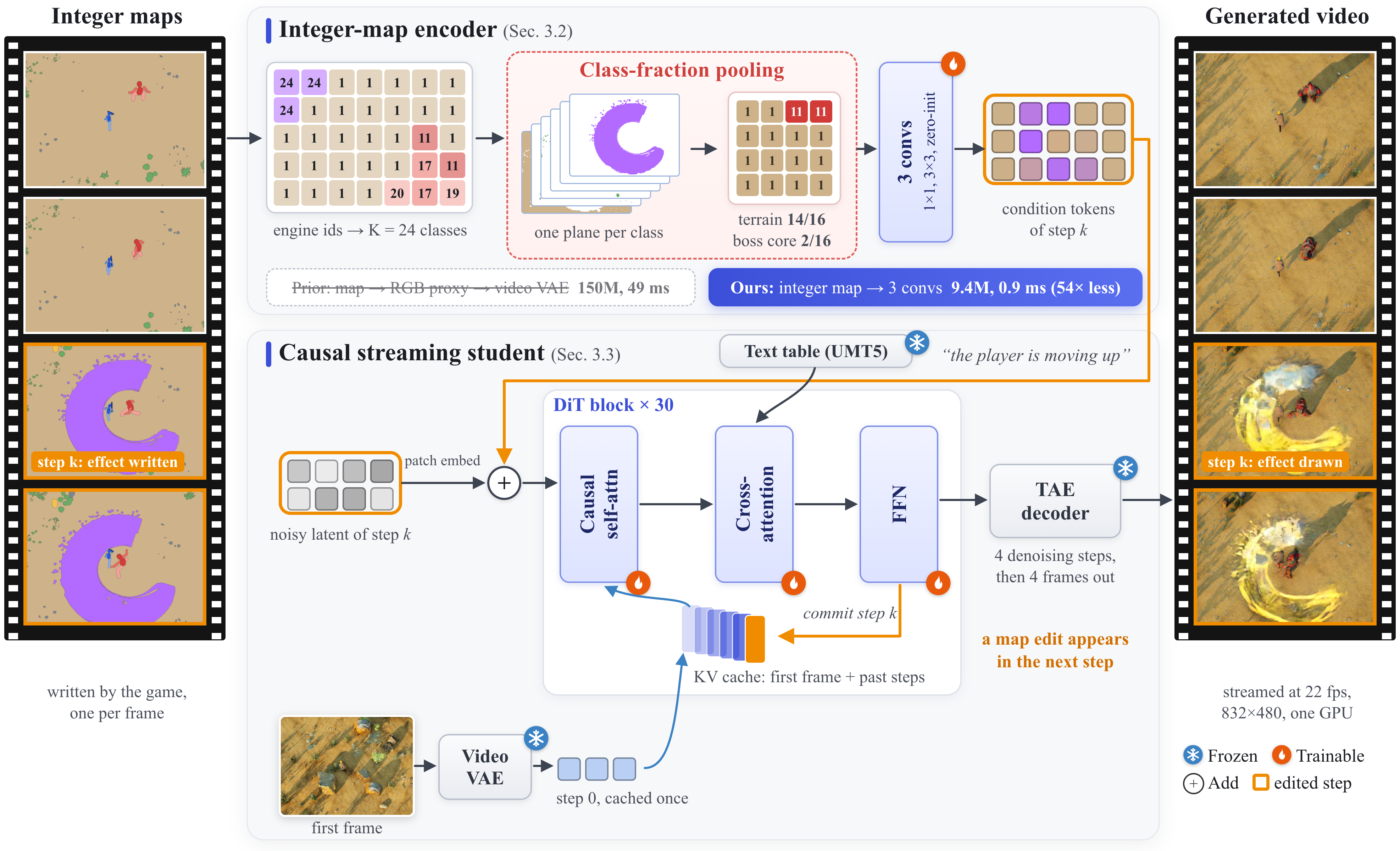}
  \caption{\textbf{Overview of MaLiang} on a held-out clip. \textbf{Encoder}: class-fraction pooling turns each $4{\times}4$
  block of the integer map into class shares (a 2-pixel boss part keeps $2/16$), and three convolutions map the four maps of
  a step to condition tokens, replacing the rendered proxy and video VAE of prior work. \textbf{Causal student}: the
  first frame is cached once; each step adds its condition tokens to the noisy latent, attends to earlier steps and the
  action text, and is committed to the cache. Orange: the step in which the game writes attack-effect classes into the
  map, traced to the frame that draws the effect.}
  \label{fig:method}
\end{figure*}

\section{Related Work}
\label{sec:related}

\paragraph{Neural game engines and world models.} Action-conditioned video models can simulate a game end to end
\cite{gamegan2020,gamengen2025,diamond2024,genie2024,oasis2024,gamegenx2025,gamefactory2025,matrixgame2,yume2025}. They keep the state implicitly in the generated pixels,
which makes long-term consistency and exact game logic hard. A recent alternative keeps a conventional or agent-written
game program as the source of truth and uses the video model only as a renderer
\cite{magpie2026,matrixgame4,streamrender2026,cwm2026,pwm2026}. MaLiang follows this split; our contribution is the
interface between the two halves.

\paragraph{Structure-conditioned video generation.} Semantic-map-to-image and -video synthesis has a long history
\cite{pix2pix2017,pix2pixhd2018,spade2019,vid2vid2018,wcvid2vid2020,makeascene2022}, game-engine buffers have been used to
make rendered frames photoreal \cite{playingfordata2016,epe2023}, and adapters such as ControlNet and T2I-Adapter
\cite{controlnet2023,t2iadapter2024,controlvideo2024,videocomposer2023} condition diffusion models on edge, depth or
segmentation \emph{images}. GAN-based renderers read one-hot label maps at pixel resolution; latent video diffusion
models instead encode an image proxy with the VAE, and game renderers inherit this image-shaped interface. We return
to label maps, but unlike SPADE-style modulation at every decoder resolution \cite{spade2019}, we add class fractions
once, at latent resolution and grouped in time like the video latent, so each causal step reads only its own slice of
the map; this is what lets edits stream step by step and keeps effects time-locked. SolarRenderer
\cite{solarrenderer2026} and AnyGameRenderer \cite{anygamerenderer2026} adapt a 33B reference-video model with LoRA to turn
semantic videos or grey blockouts into polished footage; StreamRender-H3 \cite{streamrender2026} streams the same model
from a flat-coloured semantic video; CWM \cite{cwm2026} feeds depth and semantics rendered from a code-maintained state;
Magpie \cite{magpie2026} and Matrix-Game~4.0 \cite{matrixgame4} condition real-time students on whitebox or low-poly
engine renders. PWM \cite{pwm2026} is closest in spirit: it rasterises 3D boxes into identity, semantic and direction
embedding maps that drive a ControlNet. MaLiang instead conditions on a dense integer class map that covers the whole
scene, including body parts and effects, through a light additive encoder rather than a visual encoder or a ControlNet
branch; a side-by-side summary of all interfaces is given in Appendix~\ref{sec:app-interfaces}.

\paragraph{Real-time video diffusion.} Latent video diffusion transformers \cite{ldm2022,dit2023,vdm2022,cogvideox2025,wan2025}
are slow to sample; few-step and causal students make interactive generation possible: consistency
distillation \cite{cm2023,lcm2023}, distribution matching distillation \cite{dmd2024,dmd2_2024}, and causal students trained
with teacher forcing or on their own rollouts \cite{diffusionforcing2024,causvid2025,selfforcing2025}. We use this recipe unchanged; what it
enables here is that each streaming step reads a new slice of the integer map.

\section{Method}
\label{sec:method}

\subsection{The integer map}
\label{sec:map}

Fig.~\ref{fig:method} contrasts our conditioning path with the visual-proxy path. For every output frame the game
produces one 832$\times$480 map of 8-bit engine ids, rasterised with the game camera (the \emph{numerical proxy}); a fixed lookup table maps the ids
to $K{=}24$ model classes (e.g.\ engine ids 22--25 are the four effect classes, 255 is \emph{unknown}).
The classes are: seven scene categories (sky, terrain, paving, architecture, rock, vegetation, prop);
six body parts (core, both arms, both legs, weapon) for the player and for the boss; four attack-effect classes (light
and heavy attacks of two boss families); and an \emph{unknown} id. Characters are drawn from articulated capsules
by ray casting; the same capsule code labels the training data and rasterises the live game, so the model sees the
same kind of map in both settings.

\subsection{Conditioning the video transformer}
\label{sec:cond}

The backbone is Wan2.2-TI2V-5B \cite{wan2025}, whose VAE compresses time by 4 and space by 16 and whose transformer
uses $2{\times}2$ patches. Let $s\in\{0,\dots,K{-}1\}^{T\times H\times W}$ be the id video. The encoder
$E$ turns it into a tensor of the patch-embedding shape without any visual encoding:
\begin{enumerate}
\item \emph{class pooling}: the one-hot encoding of $s$ is average-pooled over $4{\times}4$ pixel blocks, giving the
fraction of each class per block (computed class by class, so the full-resolution one-hot is never stored);
\item \emph{temporal grouping}: frames are grouped as in the VAE (frame 0 alone, then groups of four),
and the group and a $4{\times}4$ space-to-depth are folded into channels;
\item \emph{projection}: a $1{\times}1$ convolution, a $3{\times}3$ convolution (both with SiLU) and a zero-initialised
patch-stride convolution map the result to the model width $d{=}3072$.
\end{enumerate}
In this \emph{Native Numerical Conditioning}, the output is added to the patch embedding of the noisy latent,
$x = \mathrm{PatchEmbed}(z_t) + E(s)$. Zero
initialisation makes the conditioned model identical to the pretrained one at the start of training. All temporal
kernels have size one, so the condition of latent step $j$ depends only on the frames of step $j$; this is what allows
the causal distillation below. The encoder has 9.4M parameters (0.19\% of the backbone). Class fractions keep the composition of a small region
explicit, avoid the ambiguity of colours mixed by pooling, and are grouped in time exactly like the video latent.

Appearance is set by the first frame, given as a clean latent (image-to-video). A short action caption per four-frame
step (one of 18 templates describing the player's movement) is attended only by the tokens of its own step. During
training we drop the text with probability 0.1, replace the whole map by \emph{unknown} with probability 0.05, and
mask random regions to \emph{unknown} with probability 0.5, so the model also learns to render from partial maps.

\subsection{Real-time causal student}
\label{sec:distill}

The bidirectional teacher generates five-second clips with 50 sampling steps. We distil it into a causal student in
three stages: (i) \emph{causal adaptation} with teacher forcing, where each noisy step attends only to clean previous
steps; (ii) \emph{causal consistency distillation} \cite{cm2023} onto the four-step schedule
$t\in\{1000,750,500,250\}$; (iii) \emph{self-forcing distribution matching} \cite{selfforcing2025,dmd2_2024}, where the
student generates its own rollouts with a key-value cache (3 sink steps and the 17 most recent steps) and a frozen
bidirectional teacher and a trained critic provide the distribution-matching gradient. Every step of every rollout is
conditioned on its own slice of the real map. We deploy the EMA weights of stage (iii) and decode latents with the
TAEHV tiny autoencoder \cite{taehv}.

At run time the student generates one latent step (four frames, 0.25\,s of video) at a time from the cached history, the
current map slice and the current caption. Editing the world therefore needs no extra network and no separate editing sampler: the game writes the new
classes into the map, and the next step is generated from it.

\section{Experiments}
\label{sec:exp}

\begin{table*}[t]
\centering
\caption{\textbf{Main comparison} on 100 held-out clips, one protocol for all methods (baselines off the shelf). Warp:
warping error ($\times10^{-3}$, ground-truth flow). \ns: not significantly different from ours (all other differences are,
over clips and recordings). FPS at each method's native resolution on the same GPU type; baselines run in their whole-clip configuration.}
\label{tab:main}
\small
\setlength{\tabcolsep}{4pt}
\resizebox{\textwidth}{!}{%
\begin{tabular}{lccccccccccc}
\toprule
& & \multicolumn{4}{c}{Appearance} & Layout & \multicolumn{3}{c}{Entities (player, boss)} & Temporal & \\
\cmidrule(lr){3-6}\cmidrule(lr){7-7}\cmidrule(lr){8-10}\cmidrule(lr){11-11}
Method & Backbone & PSNR$\uparrow$ & SSIM$\uparrow$ & LPIPS$\downarrow$ & FVD$\downarrow$ & mIoU$\uparrow$ &
Time$\uparrow$ & Pos.$\uparrow$ & Halluc.$\downarrow$ & Warp$\downarrow$ & FPS$\uparrow$ \\
\midrule
AnyGameRenderer \cite{anygamerenderer2026} & H3 33B & 10.43 & 0.319 & 0.796 & 879 & 19.8 & 0.571 & 0.385 & 39.4 & 21.35 & 0.11 \\
CWM \cite{cwm2026} & H3 33B & 12.69 & 0.378 & 0.705 & 708 & 28.3 & 0.665 & 0.396 & 25.7 & 18.67 & 0.07 \\
SolarRenderer \cite{solarrenderer2026} & H3 33B & 18.41 & \textbf{0.533} & 0.407 & 391 & \textbf{62.7}\ns & 0.836 &
\textbf{0.598}\ns & 15.5\ns & 10.70\ns & 0.75 \\
\textbf{MaLiang (ours)} & Wan2.2 5B & \textbf{20.63} & 0.508 & \textbf{0.327} & \textbf{154} & 62.3 & \textbf{0.893} &
0.586 & \textbf{12.2} & \textbf{10.33} & \textbf{21.8} \\
\midrule
\textcolor{gray}{Ground-truth video} & -- & -- & -- & -- & -- & \textcolor{gray}{77.4} & -- & -- & -- &
\textcolor{gray}{4.57} & -- \\
\bottomrule
\end{tabular}}
\end{table*}
\begin{figure*}[t]
  \centering
  \includegraphics[width=\linewidth]{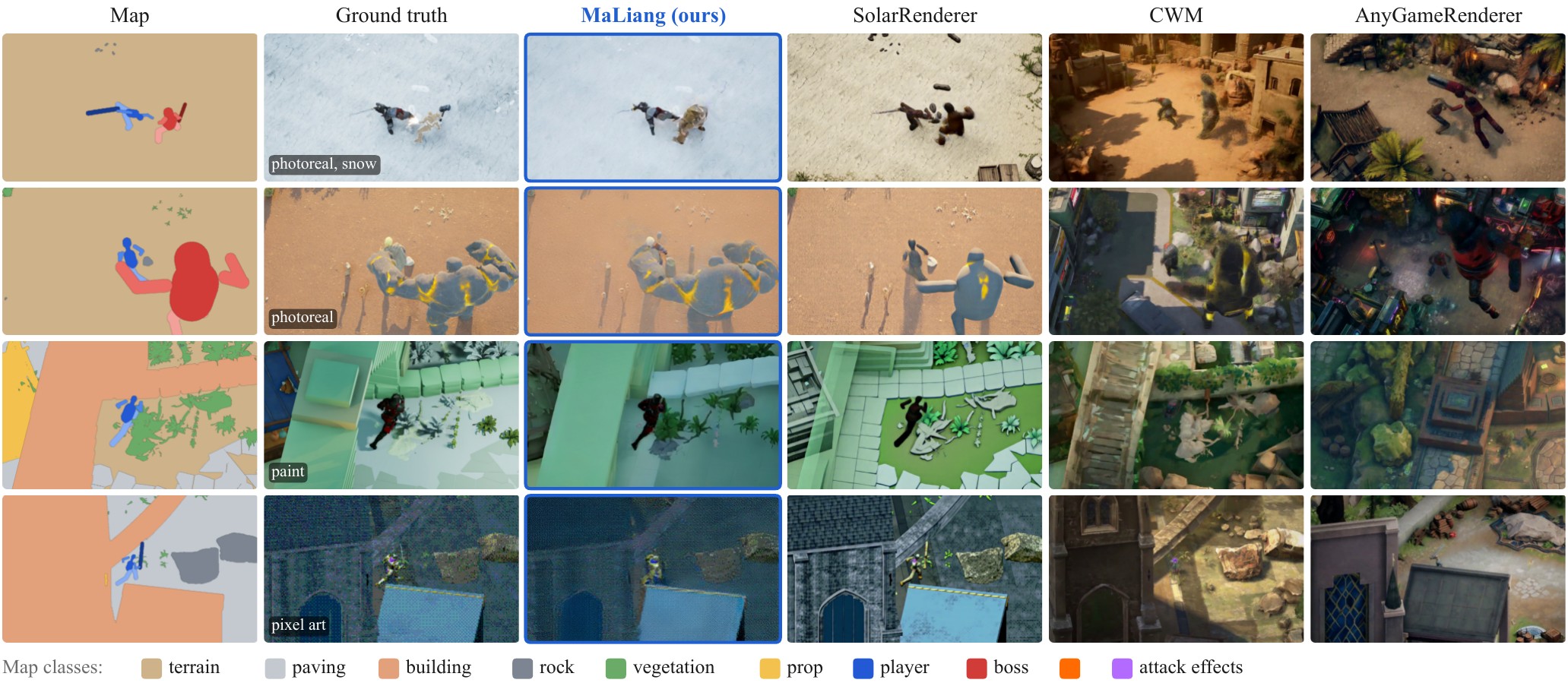}
  \caption{\textbf{Qualitative comparison} on held-out clips (one frame each). MaLiang follows the map and keeps the look of
  the recording, from snow and the golem's glowing cracks to painted and pixel-art styles; SolarRenderer keeps the layout but
  loses the characters' detail and the style, and CWM and AnyGameRenderer drift to a different scene. All 100 clips: Tab.~\ref{tab:main}.}
  \label{fig:qual}
\end{figure*}
\subsection{Setup}
\label{sec:setup}

\paragraph{Data.} Our study covers one genre: a top-down action game with a player and a boss, recorded in ten Unreal
Engine scenes (256 recordings). Clips are 5\,s (81 frames) at 16\,fps and 832$\times$480, each with RGB frames, integer map,
camera matrices and character capsules. The training set has 5,205 photoreal clips with a perspective camera and
6,750 re-styled clips with an orthographic camera. Boss attack effects from a sprite library are composited
onto 1,893 photoreal training clips and labelled in the map with the four effect classes. The teacher is trained for
10,667 steps at batch 12 (learning rate $10^{-5}$, encoder $10^{-4}$); distillation takes 8,000, 8,000 and
500 steps at batch 8.

\paragraph{Test set.} 100 clips from ten recordings that are excluded from training (ten clips each; 50 photoreal and 50
stylised; all ten scenes). The scenes themselves appear in training through other recordings; worlds that never
appear in training are evaluated in Sec.~\ref{sec:general}. Training and test are split by recording: no
recording contributes clips to both. All methods are run on the same clips and receive the
conditions through their native interfaces (see Baselines).

\paragraph{Baselines.} We ran three structure-to-video renderers off the shelf, with their public code and weights and
without fine-tuning on our game, each with the input format it was trained on, derived from our maps: \textbf{SolarRenderer} \cite{solarrenderer2026} (colour-coded semantic
video in its palette, first frame, text), \textbf{AnyGameRenderer} \cite{anygamerenderer2026} (grey geometry with
colour-coded player, boss and effects, plus entity descriptions; it takes no first frame), and \textbf{CWM}
\cite{cwm2026} (first frame, per-frame depth and semantic ids). Our data has no depth, so CWM's depth is estimated from
the whitebox frames with Depth-Anything-V2 \cite{depthanything2_2024} and scaled with the recorded camera's ground plane.
All three use the 33B MiniMax-H3 backbone \cite{minimaxh3}; we run each in its whole-clip configuration
(SolarRenderer also offers a block-causal real-time variant, which we did not use). Matrix-Game~4.0 \cite{matrixgame4} renders only its own engine scenes, and StreamRender-H3, Magpie and PWM have
no runnable release for our setting; they are compared by interface in Appendix~\ref{sec:app-interfaces}.

\paragraph{Protocol and metrics.} Every output is resampled to 832$\times$480 at 16\,fps (24-fps outputs take frame
$\lceil 1.5k\rceil$ for source frame $k$), and all metrics use frames 1--80 (frame 0 is given). \emph{Appearance}: PSNR,
SSIM, LPIPS \cite{lpips2018} and FVD \cite{fvd2018} (I3D, five 16-frame windows per clip). \emph{Layout control}: mIoU
between a SegFormer-B2 \cite{segformer2021} segmenter, fine-tuned on recorded frames and their maps (9 classes; 77.4\%
on the ground-truth frames, the segmenter's reference value; trained on frames of 2,400 training clips), and the input map. \emph{Entity control}, following
\cite{anygamerenderer2026}: time match (a character is rendered exactly when the map contains it), position IoU and
hallucination rate, averaged over player and boss. \emph{Temporal stability}: the warping error of \cite{lai2018}, using
RAFT \cite{raft2020} flow and occlusion masks of the ground-truth video. \emph{Significance}: paired bootstrap over clips
(2,000 resamples; 1,000 for FVD) and over whole recordings; we call a difference significant only
if both 95\% intervals exclude zero.

\begin{figure*}[t]
  \centering
  \includegraphics[width=\linewidth]{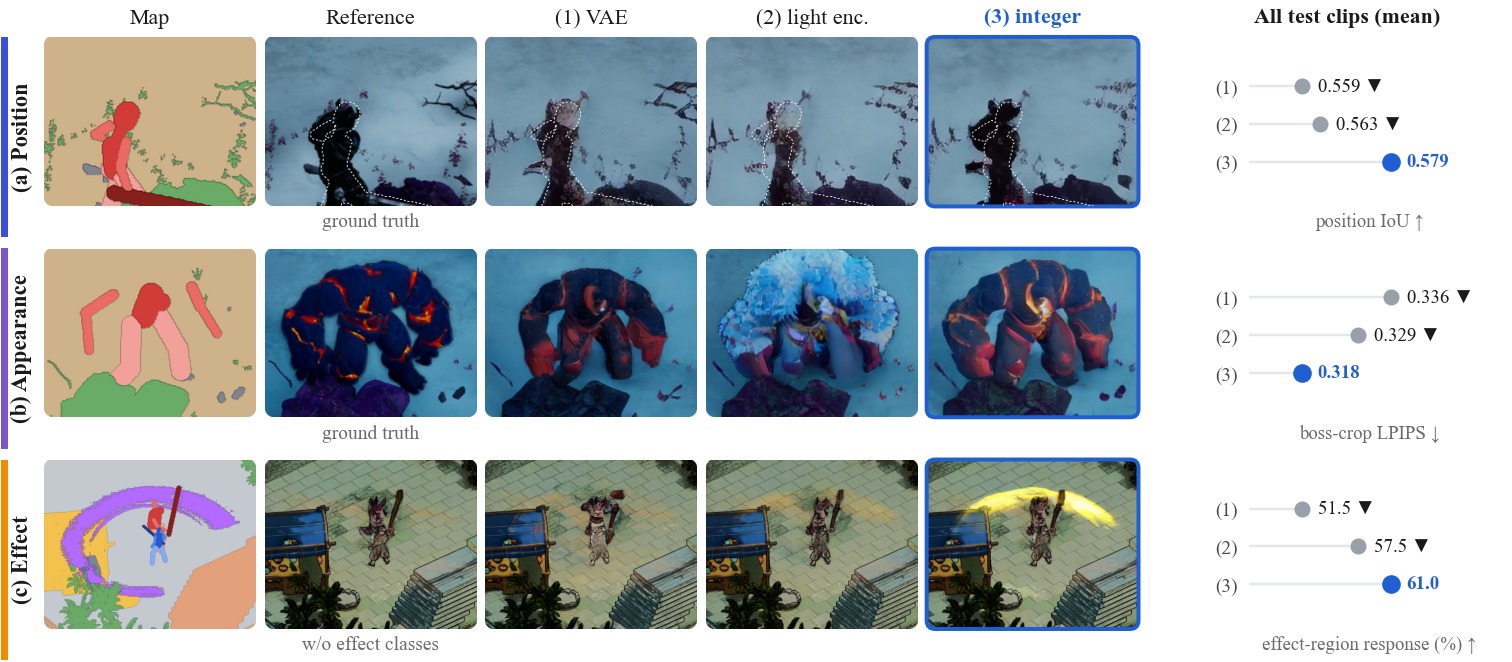}
  \caption{\textbf{Same backbone, different interface} (Tab.~\ref{tab:ablation} teachers, same seed and first frame).
  Left: one held-out frame per aspect; (a) white outline: the character's mask in the map, (c) an attack effect (purple)
  added to the map of a recording that has none; this case has no ground truth, so Reference is variant (3) on the same map
  without the effect classes. Right: mean over all test clips ($\blacktriangledown$: significantly worse
  than (3) over clips and recordings); boss-crop LPIPS is computed on the 55 clips that show the boss. More examples in
  Appendix~\ref{sec:app-abl}.}
  \label{fig:ablcrops}
\end{figure*}
\begin{table*}[t]
\centering
\caption{\textbf{Interface ablation}: same backbone, data and 5,000 steps, 50-step bidirectional sampling, 100 test clips;
effects: 94 boss-attack clips (Resp.: effect-region response, a control metric rather than a quality metric: the share of effect pixels whose mean absolute difference
to the run without effect classes exceeds 25 grey levels; Timing: share of frames where the effect appears exactly when
mapped).
$^{\blacktriangle}$/$^{\blacktriangledown}$: significantly better/worse than (3) under both bootstrap schemes. Bottom:
(2) and (3) distilled into 4-step causal students with one budget (2,000 + 2,000 steps), marks relative to student (3).}
\label{tab:ablation}
\small
\setlength{\tabcolsep}{3.6pt}
\resizebox{\textwidth}{!}{%
\begin{tabular}{lcccccccccccc}
\toprule
& & \multicolumn{4}{c}{Appearance} & Layout & \multicolumn{3}{c}{Entities} & Temporal & \multicolumn{2}{c}{Attack effects} \\
\cmidrule(lr){3-6}\cmidrule(lr){7-7}\cmidrule(lr){8-10}\cmidrule(lr){11-11}\cmidrule(lr){12-13}
Condition path & Enc.\,ms & PSNR$\uparrow$ & SSIM$\uparrow$ & LPIPS$\downarrow$ & FVD$\downarrow$ & mIoU$\uparrow$ &
Time$\uparrow$ & Pos.$\uparrow$ & Halluc.$\downarrow$ & Warp$\downarrow$ & Resp.$\uparrow$ & Timing$\uparrow$ \\
\midrule
\multicolumn{13}{l}{\emph{50-step bidirectional teachers}} \\
(1) colour $\to$ video VAE & 49.4 & 20.55 & 0.502 & 0.306 & 110 & 65.4 & 0.863$^{\blacktriangledown}$ &
0.559$^{\blacktriangledown}$ & 12.3 & 9.90$^{\blacktriangledown}$ & 51.5$^{\blacktriangledown}$ & 87.1$^{\blacktriangledown}$ \\
(2) colour $\to$ light enc. & $\approx$(3)$^{\ast}$ & 20.57 & 0.518$^{\blacktriangle}$ &
0.294$^{\blacktriangle}$ & 114 & 66.7 & 0.871 & 0.563$^{\blacktriangledown}$ & 11.7 & 8.78 & 57.5$^{\blacktriangledown}$ & 90.8$^{\blacktriangledown}$ \\
\textbf{(3) integer (ours)} & 0.9 & 20.37 & 0.509 & 0.301 & 119 & 65.8 & \textbf{0.880} & \textbf{0.579} & 11.5 &
8.89 & \textbf{61.0} & \textbf{93.7} \\
\quad (3a) one id per $4{\times}4$ block & 0.9 & 20.66$^{\blacktriangle}$ & 0.509 & 0.301 & 117 & 65.4 & 0.872$^{\blacktriangledown}$ & 0.574$^{\blacktriangledown}$ & 10.3$^{\blacktriangle}$ & 9.13 & 57.1$^{\blacktriangledown}$ & 91.6$^{\blacktriangledown}$ \\
\quad (3b) no region masking & 0.9 & 20.73$^{\blacktriangle}$ & 0.513 & 0.298 & 118 & 66.6 & 0.873 & 0.576 & 11.1 & 9.10 & 54.9$^{\blacktriangledown}$ & 89.7$^{\blacktriangledown}$ \\
(4) no map & -- & 16.18$^{\blacktriangledown}$ & 0.399$^{\blacktriangledown}$ & 0.508$^{\blacktriangledown}$ &
634$^{\blacktriangledown}$ & 29.0$^{\blacktriangledown}$ & 0.551$^{\blacktriangledown}$ & 0.376$^{\blacktriangledown}$ &
23.9$^{\blacktriangledown}$ & 26.67$^{\blacktriangledown}$ & -- & -- \\
\midrule
\multicolumn{13}{l}{\emph{4-step causal students, same distillation budget}} \\
(2) colour $\to$ light enc. & $\approx$(3)$^{\ast}$ & 18.65 & 0.463$^{\blacktriangle}$ & 0.401$^{\blacktriangledown}$ & 290 & 51.4$^{\blacktriangledown}$ & 0.675$^{\blacktriangledown}$ & 0.523$^{\blacktriangledown}$ & 16.7 & 11.99 & 49.3$^{\blacktriangledown}$ & 83.7$^{\blacktriangledown}$ \\
\textbf{(3) integer (ours)} & 0.9 & 18.64 & 0.455 & \textbf{0.388} & 298 & \textbf{53.4} & \textbf{0.711} & \textbf{0.541} & 16.5 & 12.65 & \textbf{58.1} & \textbf{87.5} \\
\bottomrule
\end{tabular}}\\[2pt]
{\footnotesize Encoder time per four-frame step on one idle GPU; for (1), the VAE encode of an 81-frame clip amortised over its 21 steps (parameters in Tabs.~\ref{tab:app-config} and~\ref{tab:app-cost}: 150M VAE, 8.8M / 9.4M light encoders). $^{\ast}$Same architecture as (3); timed side by side with (3) on a shared GPU: 2.96 vs.\ 3.08\,ms.}
\end{table*}
\subsection{Comparison with structure-to-video renderers}
\label{sec:main}

Tab.~\ref{tab:main} is a system-level comparison: the methods differ in backbone, training data and resolution, and
the interface itself is isolated in Sec.~\ref{sec:ablation}. MaLiang has the best FVD, LPIPS and PSNR by a wide margin: FVD
falls from 391 to 154 and LPIPS from 0.407 to 0.327 relative to SolarRenderer, and PSNR rises by 2.2\,dB (95\% CI
[1.8, 2.7]). SolarRenderer keeps a slightly higher SSIM (0.533 vs.\ 0.508). Layout control is statistically tied (mIoU 62.3 vs.\ 62.7, $p{=}0.88$), as are position IoU and temporal
warping error; MaLiang renders characters significantly more often when and only when they are present (time match
0.893 vs.\ 0.836). CWM and AnyGameRenderer lose the layout within a few seconds:
AnyGameRenderer receives no first frame, so its appearance comes from text alone, and CWM matches the first frame (about
31\,dB) but drifts into a different layout within two seconds. All three baselines use a 33B backbone and, in the whole-clip configurations we ran, a
30--310$\times$ lower frame rate (11--120$\times$ fewer pixels per second at 1344$\times$768; generation only, e.g.\ a
median of 211\,s per 158-frame clip for SolarRenderer); a clip is shown only once it is complete. Real-time renderers with a visual proxy are discussed in Sec.~\ref{sec:costexp}.

\begin{figure*}[t]
  \centering
  \includegraphics[width=\linewidth]{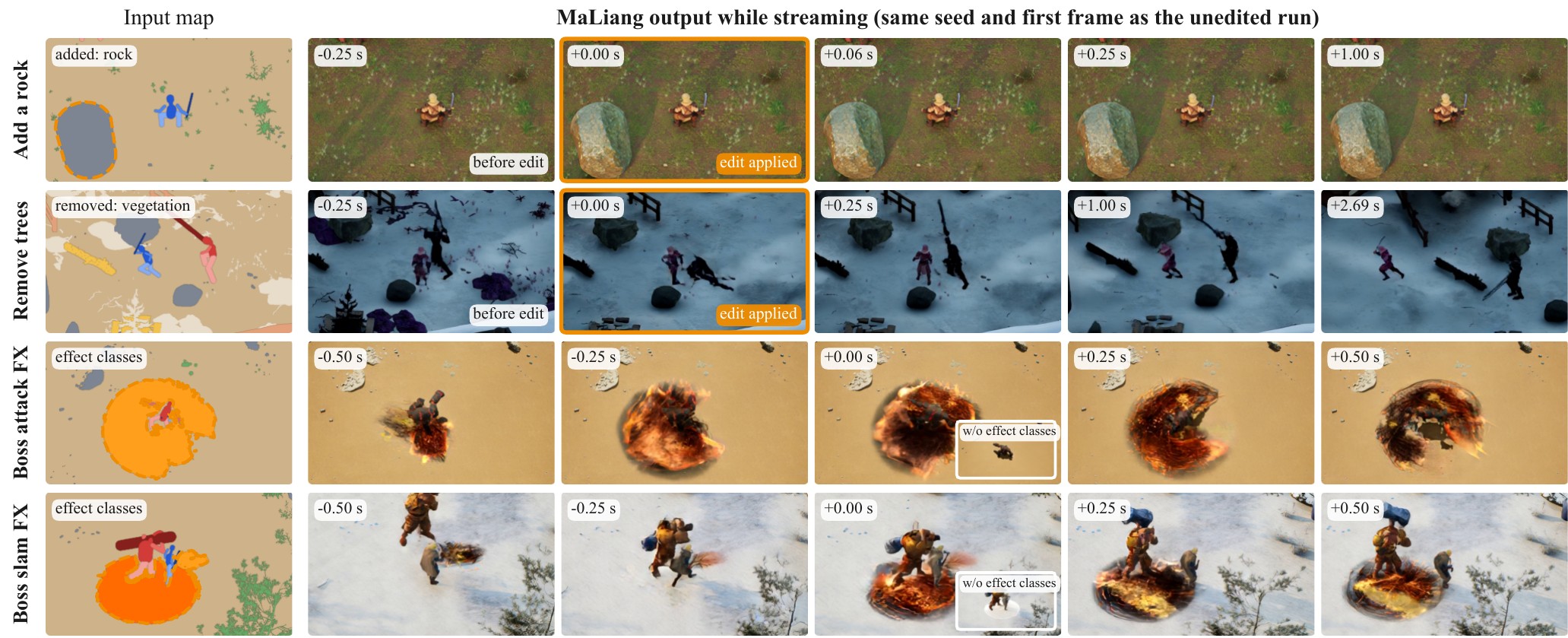}
  \caption{\textbf{Live control while streaming.} Times are relative to the edit. Rows 1--2: an object is written into, or a
  category removed from, the map at $t{=}0$ (orange: the step the edit is applied); the change appears in the next
  generated step. Rows 3--4: attack-effect classes in the map are rendered as boss attack effects, time-locked to the
  map (times relative to the frame where the effect region in the map is largest); inset: the same seed and map without the effect classes. All frames are outputs of the deploy student; Sec.~\ref{sec:edit}
  reports all 33 edits and 94 clips.}
  \label{fig:control}
\end{figure*}

\subsection{Is it the interface? A same-backbone ablation}
\label{sec:ablation}

To isolate the interface, we train four variants from the same
Wan2.2-TI2V-5B weights on the same data with the same recipe for 5,000 steps, and sample all of them as bidirectional
teachers with 50 steps: \textbf{(1)} the map colour-coded into an RGB video and encoded by the frozen video VAE, added
through a zero-initialised projection after the patch embedding; \textbf{(2)} the same colour video through a light
encoder with the architecture of ours (per-block mean and standard deviation of the colour instead of class
fractions); \textbf{(3)} integer
conditioning (ours); \textbf{(4)} no map (all ids unknown).

Tab.~\ref{tab:ablation} shows four things. First, the map matters: without it every metric collapses (mIoU 29.0, FVD 634).
Second, layout control and image quality match colour images: PSNR, FVD, mIoU and
hallucination rate are tied; (2) is slightly but significantly better on SSIM and LPIPS, by less than 0.01. Third, integer conditioning localises characters better: position IoU is
significantly higher than with either colour interface ($+0.020$ over (1), $+0.016$ over (2); Fig.~\ref{fig:ablcrops}a), and against the VAE path it
also improves time match and temporal stability; the boss itself is also rendered more faithfully (boss-crop LPIPS 0.318 vs.\ 0.336 and 0.329, significant), whereas with
colour codes a boss can even change material (Fig.~\ref{fig:ablcrops}b). Fourth, integer conditioning controls attack effects more reliably.
In the effect test of Sec.~\ref{sec:edit} (same 94 clips, seed and first frame), the effect-region response is 61.0\% with integers, against 51.5\% with the
VAE path and 57.5\% with the light colour encoder ($+9.5$ [6.9, 12.4] and $+3.5$ [1.6, 5.3] points, clip bootstrap; also over recordings), and the effect appears exactly when the map contains it in 93.7\% of frames
against 87.1\% and 90.8\% (Fig.~\ref{fig:ablcrops}c). Changes outside the effect region are equally rare for all three (about 1\%). After $4{\times}4$ pooling a colour code averages a thin limb or a short-lived effect with its surroundings, whereas
class fractions keep its share (Fig.~\ref{fig:method}). Both encoder choices matter: one id per block (3a) or no region
masking (3b) lowers the effect-region response by 3.9 / 6.1 points and correct timing by 2.1 / 4.0 points. Against the VAE path,
removing the visual encoder cuts the condition cost about 50-fold; against the equally cheap light colour encoder (2),
the integer form adds better control of characters and effects: the advantage is precision, not speed.

\begin{figure*}[t]
  \centering
  \includegraphics[width=\linewidth]{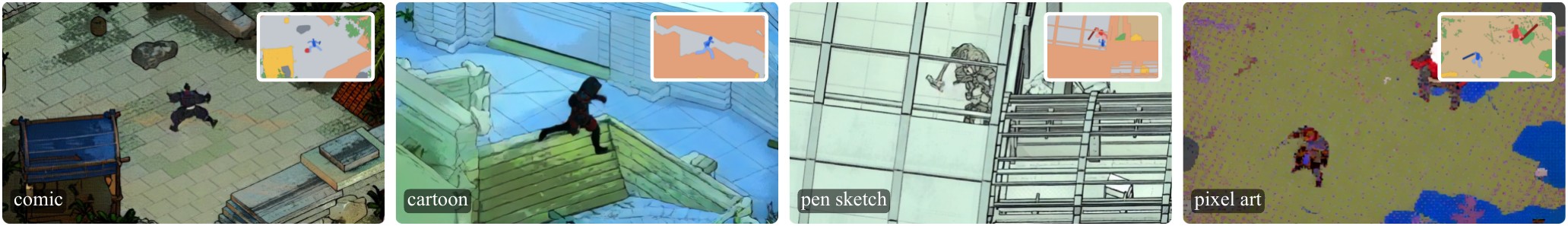}
  \caption{\textbf{Stylised worlds}: real-time student on held-out comic, cartoon, pen-sketch and pixel-art recordings
  (inset: input map).}
  \label{fig:styles}
\end{figure*}
\paragraph{Does the advantage survive distillation?} The system runs a 4-step causal student, so we distilled (2) and (3)
with one reduced budget (2,000 teacher-forcing + 2,000 consistency steps, no DMD) and ran both with the streaming
sampler (Tab.~\ref{tab:ablation}, bottom). Under this matched budget the gap widens: the integer student is significantly better in mIoU
($+2.0$), time match ($+0.036$), position IoU ($+0.018$), LPIPS ($-0.013$) and in effect control ($+8.8$ points of
effect-region response, $+3.8$ points of correctly timed frames), while PSNR, FVD, hallucination and warping error are tied and SSIM slightly favours (2). The integer interface is thus also more robust to distillation. The deployed student (Tab.~\ref{tab:main}), trained longer and with DMD, beats both.

\subsection{Live editing and effect control}
\label{sec:edit}

\paragraph{Edits.} Each edit is defined in world coordinates and re-projected every frame with the recorded camera: an
upright cylinder of rock, vegetation or architecture placed on visible ground away from the characters, or the removal of
the largest static category in view (refilled with the surrounding ground). We run the student twice with the same seed,
first frame and captions, once on the original and once on the edited map, and segment the outputs (Fig.~\ref{fig:control}). Edited pixels take
the new class in 69.1\% of cases, against 13.6\% for the same pixels without the edit; adding and removing gain almost
the same ($+56.4$ and $+54.2$ points). An edit is visible 1.3 frames after it is
applied, while the rest of the frame stays close to the unedited run (31.5\,dB PSNR).

\paragraph{Effects.} Boss attack effects are four classes of the map, so a game can switch them on per pixel. On 94
held-out clips with boss attacks, we run the student with and without the effect classes in the map (same seed,
otherwise identical maps). Inside the effect region 63.3\% of pixels change (mean absolute difference above 25 grey levels), outside it (beyond a 49-px margin) only 1.3\%, and in 93.8\% of
frames the effect is visible exactly when the map contains it (Fig.~\ref{fig:control}, rows 3--4). The test effects come from a second sprite library, so the map controls effect shapes and trajectories that do not
occur in training. No baseline has effect classes: SolarRenderer and AnyGameRenderer receive effects only as
colour-coded regions, and CWM drops them.

\paragraph{Interactive demo.} We also built a browser game that rasterises its state into the map every 0.25\,s and
streams the student (Fig.~\ref{fig:teaser}b; Appendix~\ref{sec:app-demo}). The map is a \emph{Numerical Canvas}: gameplay rules only write classes into it (a
boss clone, walls, a tower, falling rocks, vegetation, a giant boss, effects); these rules never occur in training, yet the unchanged student renders each from the map alone.

\subsection{Cost and latency}
\label{sec:costexp}

The integer encoder takes 0.91\,ms per step, 54$\times$ less than the video-VAE encode of a colour proxy (49.4\,ms per
step, amortised over an 81-frame clip), and is a
negligible part of the 184-ms step (transformer plus decoding; 95th percentile
192\,ms; full breakdown in Tab.~\ref{tab:app-cost}). A light colour encoder or a tiny autoencoder is equally cheap (estimated 21.7 vs.\ 21.8\,fps); the gain of integers is
control precision (Sec.~\ref{sec:ablation}), and the proxy must still be rendered and carries twice the data. With one step per 0.25\,s of video, the student runs at 1.4$\times$ real time. Two latencies
should be kept apart: in the generated video an edit is visible 1.3 frames after the frame where it is applied
(Sec.~\ref{sec:edit}); in wall-clock time a change written into the map reaches the screen within two steps (median
0.28\,s, 95th percentile 0.37\,s, estimated for a change arriving at a uniformly random time within a step; network
transport not included).

\paragraph{Real-time proxy renderers.} Magpie reports 32\,fps at 1280$\times$768 on an H100 with three steps
\cite{magpie2026}, and Matrix-Game~4.0 60\,fps at 480p on an RTX~5090 with one step \cite{matrixgame4}; with its
released real-time preset we measured 45--54\,fps on an idle GPU of our type, about twice our 21.8\,fps. It uses a 5B Wan2.2
transformer like ours, so the gap comes mainly from one denoising step instead of four, which is orthogonal to the
interface. Both systems still render an engine proxy every frame (Magpie: 0.83\,s of engine preparation in its 1.55-s
first-feedback latency).

\subsection{Unseen worlds and long horizons}
\label{sec:general}

\paragraph{Unseen worlds.} Procedural worlds (four themes) and freehand-style sketches are rasterised with test-clip cameras and
characters, with a first frame from an image model \cite{flux2}. These maps have layouts unseen in training. On canyon, forest, plaza and village maps the student reaches 56.1 / 64.9 / 39.1 / 49.5\% mIoU, against
44.1 / 50.6 / 32.5 / 49.1\% for SolarRenderer, and 38.7 vs.\ 37.2\% on sketches (Tab.~\ref{tab:app-unseen}). As the map carries layout and the first frame carries appearance, one map
can be rendered in many looks (Fig.~\ref{fig:teaser}c), and held-out recordings in stylised looks are rendered too (Fig.~\ref{fig:qual}, rows 3--4;
Fig.~\ref{fig:styles}).

\paragraph{Long horizons.} Over 60\,s streams on ten chains of held-out clips (curves: Fig.~\ref{fig:app-long}), layout control on
photoreal chains shows no decline (pooled mIoU 57.8\% in the first ten seconds, 62.3\% in the last), and stylised
chains still follow the map after one minute (39.9\%).

\section{Limitations}
\label{sec:limits}

Our study covers one top-down action game. The 24-class vocabulary is fixed, so new object types need new classes and
retraining; appearance comes entirely from the first frame, so materials and character identity change only by
re-drawing it; and the 2D map carries no depth, so height and occlusion are left to the model. Material, identity and
depth classes, and other engines that rasterise object ids, including first-person games, are natural next steps.

\section{Conclusion}
\label{sec:conclusion}

A generative game renderer need not see its input as a picture: with an integer map and a light additive encoder, a game
changes the rendered world, or switches on an effect, simply by writing a number into the map.

{
    \small
    \bibliographystyle{plainnat}
    \bibliography{refs}
}

\clearpage
\appendix
\setcounter{figure}{0}\setcounter{table}{0}
\renewcommand{\thefigure}{S\arabic{figure}}\renewcommand{\thetable}{S\arabic{table}}
\renewcommand{\theHfigure}{S\arabic{figure}}\renewcommand{\theHtable}{S\arabic{table}}
\newcommand{\up}{\ensuremath{\blacktriangle}}
\newcommand{\dn}{\ensuremath{\blacktriangledown}}

\twocolumn[{\centering\Large\bfseries Appendix\\[1.2em]}]

Image sources by figure: Fig.~\ref{fig:app-worlds} and the demo sequences are generated by the deploy student of the main text (four-step causal model, TAEHV decoder), with first frames drawn by FLUX.2 [klein]~\cite{flux2} or re-drawn by Qwen-Image-Edit~\cite{qwenimage2025} as stated in the captions; the unseen-world and held-out comparisons show the deploy student next to the
ground truth and the off-the-shelf baselines; the same-backbone examples are generated by the 50-step bidirectional
teachers of the ablation. Frames marked as input maps are rasterised maps. Examples were selected by hand to show each
capability; quantitative results over all clips are in the main text, in Sec.~\ref{sec:app-vace},
Secs.~\ref{sec:app-seeds}--\ref{sec:app-imperfect} and Tab.~\ref{tab:app-editall}. The project page (\url{https://yeyinglab.github.io/MaLiang/}) has a 1.5-min overview video (with sound) of MaLiang on new play sessions recorded in the browser demo (Sec.~\ref{sec:app-demo}): restyled and procedural worlds, players and bosses with attack effects, map edits and
attack-triggered rules, with the input map as an inset. It also shows further cases with their maps, the main tables and a 90-second continuous rollout.

\section{Configuration}
\label{sec:app-config}
Tab.~\ref{tab:app-config} lists the settings used for training, sampling and evaluation throughout the paper.

\begin{table}[t]
\centering
\caption{\textbf{Training, sampling and evaluation configuration.}}
\label{tab:app-config}
\footnotesize
\setlength{\tabcolsep}{3pt}
\begin{tabular}{p{0.32\linewidth}p{0.62\linewidth}}
\toprule
Backbone & Wan2.2-TI2V-5B, 30 blocks, width 3072; bf16 weights and activations \\
Integer encoder & $K{=}24$ classes; per-class $4{\times}4$ average pooling; frame 0 repeated 4$\times$ then groups of 4
frames; $4{\times}4$ space-to-depth; Conv $1{\times}1$ + SiLU, Conv $(1,3,3)$ + SiLU, zero-init Conv $(1,2,2)$ stride 2;
9.4M parameters \\
Light colour encoder (2) & same layout; per-block colour mean and standard deviation instead of class fractions; 8.8M \\
Teacher training & 10,667 steps, global batch 12, AdamW (wd 0.01), lr $10^{-5}$, encoder lr $10^{-4}$, warm-up 500,
grad clip 1.0; text drop 0.1, map drop 0.05, region masking 0.5 \\
Ablation teachers (1)--(5), (3a), (3b) & same recipe, 5,000 steps \\
Distillation (deploy) & (i) teacher forcing 8,000 steps, lr $10^{-5}$; (ii) consistency 8,000 steps, lr
$2{\cdot}10^{-6}$, EMA 0.99, axis $t\in\{1000,750,500,250\}$; (iii) self-forcing DMD 500 steps, lr $2{\cdot}10^{-6}$,
critic lr $10^{-6}$, 5 critic steps, EMA 0.99; batch 8; KV cache 3 sink + 17 recent steps; rollouts of 21--27 steps \\
Distillation (Tab.~\ref{tab:ablation}, bottom) & (i) 2,000 + (ii) 2,000 steps, batch 6, no DMD; same for (2) and (3) \\
Sampling & teachers: 50 steps, bidirectional, seed 0; student: 4 steps per 4-frame step, TAEHV decoder \\
Evaluation labels & 24 model classes $\to$ 9: the seven scene classes kept; player parts $\to$ player; boss parts $\to$
boss; effect classes and \emph{unknown} ignored \\
Segmenter & SegFormer-B2 fine-tuned 12k steps on frames 0/27/54/80 of 2,400 training clips \\
FVD & I3D features of five 16-frame windows per clip (frames 1, 17, 33, 49, 65); all 500 windows of a method against
the 500 ground-truth windows \\
Significance & paired bootstrap, 2,000 resamples (1,000 for FVD), over clips and over recordings; both must exclude 0 \\
Split & by recording; no recording contributes clips to both training and test \\
\bottomrule
\end{tabular}
\end{table}

\section{Interfaces of generative game renderers}
\label{sec:app-interfaces}
Tab.~\ref{tab:app-capability} compares what each renderer discussed in the main text receives from the game.

\begin{table*}[t]
\centering
\caption{\textbf{Conditioning interfaces of generative game renderers.} ``Render pass'': the game state is drawn into a
proxy video before the video model sees it. Throughput as reported by each paper (hardware differs) or, for the methods
we ran, measured on our test set; --: not reported. PWM rasterises boxes into learned embedding maps rather than an RGB
video.}
\label{tab:app-capability}
\small
\setlength{\tabcolsep}{4pt}
\resizebox{\textwidth}{!}{%
\begin{tabular}{llcllr}
\toprule
Method & What the video model sees & Render pass & Attack effects & Streaming & Throughput \\
\midrule
Magpie & whitebox engine frames (cross-attention) & \cmark & not addressed & \cmark & 32\,fps, H100, 1280$\times$768 \\
Matrix-Game 4.0 & low-poly proxy + caption + reference frames & \cmark & as caption text & \cmark & 60\,fps, RTX 5090; 45--54\,fps, our GPU \\
StreamRender-H3 & flat-colour semantic video (13 classes) & \cmark & not addressed & \cmark & 218\,ms/step, 8$\times$GB200 \\
PWM & box identity / semantic / direction maps (ControlNet) & partial & not addressed & -- & -- \\
SolarRenderer & colour-coded semantic video (reference path) & \cmark & generic colour & \cmark$^{\dagger}$ & 0.75\,fps, our GPU, whole-clip \\
AnyGameRenderer & grey geometry + colour-coded entities + text & \cmark & colour + text & \xmark & 0.11\,fps, our GPU \\
CWM & first frame + depth + semantic video & \cmark & dropped & \xmark & 0.07\,fps, our GPU \\
\midrule
\textbf{MaLiang} & \textbf{integer class map, additive encoder} & \xmark & \textbf{native classes} & \cmark & 22\,fps, one GPU, 832$\times$480 \\
\bottomrule
\end{tabular}}
{\footnotesize $^{\dagger}$SolarRenderer offers a block-causal real-time variant; we ran its whole-clip variant.}
\end{table*}

\section{A ControlNet-style variant: the VACE branch}
\label{sec:app-vace}
ControlNet-style conditioning adds a trainable copy of part of the backbone that reads the condition and injects
features into the main branch. For our backbone the native form is Wan's VACE branch~\cite{vace2025}, which we add to the interface
ablation as variant (5): copies of the 15 even blocks ($0, 2, \ldots, 28$; about 2.4B parameters) initialised from the
backbone read the colour-coded map of (1), encoded by the frozen video VAE, and their outputs are added after the
matching main blocks through zero-initialised projections. It is trained on the same data with the same recipe as
(1)--(3) (5,000 steps, Tab.~\ref{tab:app-config}) and evaluated with the same protocol and the three seeds of
Sec.~\ref{sec:app-seeds}. It keeps both costs that Sec.~\ref{sec:ablation} isolates, the colour render and the VAE encode of (1), and
adds a second transformer of half the backbone's depth at every denoising step, roughly 0.5$\times$ the backbone's
compute.

Tab.~\ref{tab:app-vace} compares it with the integer encoder (3). The extra branch buys image quality: (5) has the
lowest FVD of all variants (105 vs.\ 116, significant in two of three seeds) and slightly better LPIPS and mIoU in one
seed each. It does not close the gap on what integers are for. Position IoU stays higher with integers in every seed
(+0.008 to +0.013), and so do the effect-region response (+7.4 to +8.7 points) and correct timing (+5.3 to +6.8); on
effects (5) is no better than the VAE path (1) (Tab.~\ref{tab:app-seeds}). The branch sees the map only after the VAE
has compressed it 16$\times$ in each spatial dimension, so a thin limb or a short-lived effect is still averaged with its
surroundings. The branch also flickers more (warping error 10.1 vs.\ 8.8, every seed). It changes the frame less outside
the effect regions (0.7 vs.\ 1.1\%), in line with its weaker response to effect classes. Integer conditioning reaches
better control at 0.9\,ms per step with 9.4M parameters, without a second branch.

\begin{table*}[t]
\centering
\caption{\textbf{ControlNet-style VACE branch (5) vs.\ integer encoder (3)} (teachers, same backbone, data and 5,000
steps; 100 test clips and the 94-clip effect test; mean $\pm$ standard deviation over seeds 0, 1, 2). Leak.: change
outside the effect regions. Bottom: number of seeds in which (3) is significantly better (\up) or worse (\dn) than (5)
under the significance rule of Sec.~\ref{sec:setup} (clip and recording bootstrap both exclude 0, tested per seed). Rows (1) and (2) are in
Tab.~\ref{tab:app-seeds}.}
\label{tab:app-vace}
\setlength{\tabcolsep}{3pt}
\resizebox{\textwidth}{!}{%
\begin{tabular}{lcccccccccccc}
\toprule
 & PSNR$\uparrow$ & SSIM$\uparrow$ & LPIPS$\downarrow$ & FVD$\downarrow$ & mIoU$\uparrow$ & Time$\uparrow$ & Pos.$\uparrow$
 & Halluc.$\downarrow$ & Warp$\downarrow$ & Resp.$\uparrow$ & Timing$\uparrow$ & Leak.$\downarrow$ \\
\midrule
(5) colour $\to$ VAE $\to$ VACE branch & 20.68$\pm$0.06 & 0.505$\pm$0.004 & 0.292$\pm$0.003 & \textbf{105}$\pm$5
 & 67.0$\pm$0.3 & 0.875$\pm$0.003 & 0.570$\pm$0.005 & 12.1$\pm$0.3 & 10.08$\pm$0.40 & 53.5$\pm$1.7 & 88.0$\pm$1.4
 & \textbf{0.74}$\pm$0.03 \\
\textbf{(3) integer (ours)} & 20.53$\pm$0.14 & 0.513$\pm$0.004 & 0.299$\pm$0.002 & 116$\pm$2 & 65.7$\pm$0.3
 & 0.878$\pm$0.003 & \textbf{0.580}$\pm$0.002 & 11.8$\pm$0.3 & \textbf{8.80}$\pm$0.27 & \textbf{61.4}$\pm$1.3
 & \textbf{94.2}$\pm$0.7 & 1.14$\pm$0.01 \\
\midrule
(3) vs.\ (5), seeds significant & 0 & 1\up & 1\dn & 2\dn & 1\dn & 0 & \textbf{3\up} & 0 & \textbf{3\up} & \textbf{3\up}
 & \textbf{3\up} & 3\dn \\
\bottomrule
\end{tabular}}
\end{table*}

\section{One map, many worlds}
\label{sec:app-worlds}
\begin{figure*}[t]
  \centering
  \includegraphics[width=\linewidth]{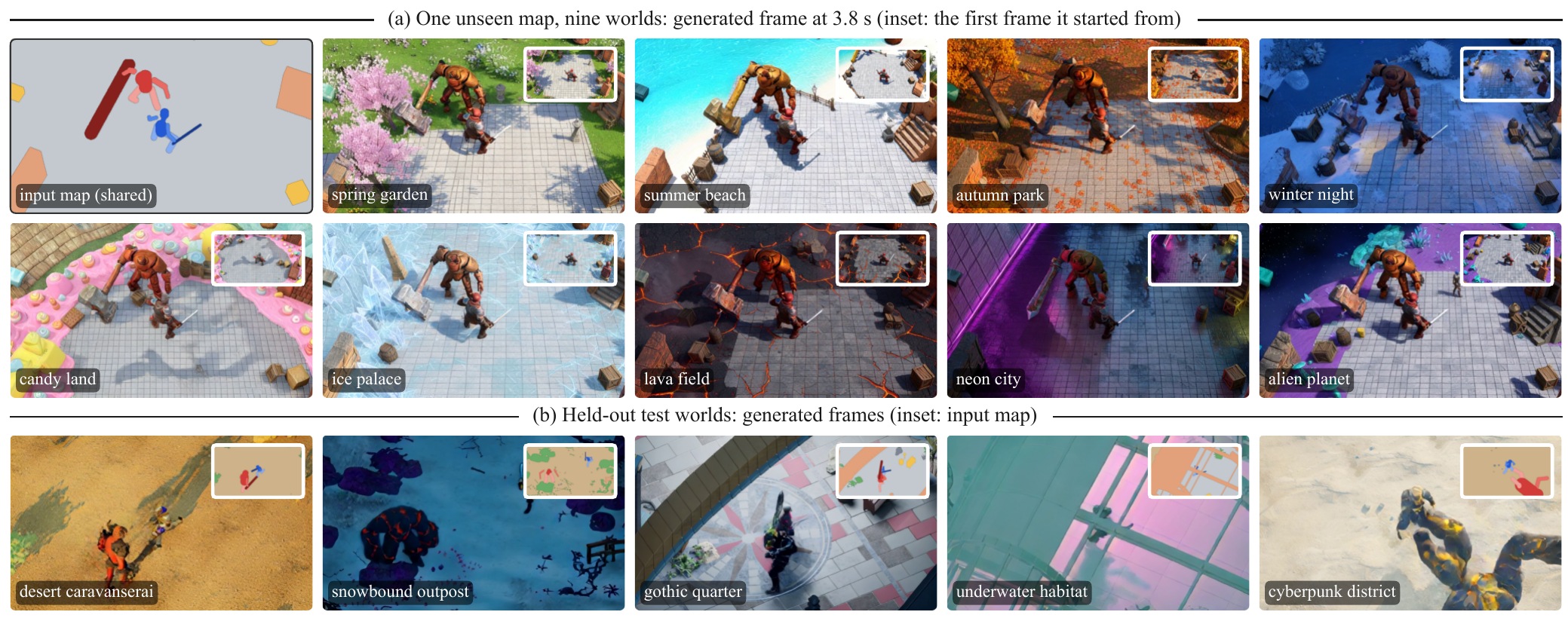}
  \caption{\textbf{Many worlds.} (a)~One unseen procedural map, nine first frames drawn by FLUX.2 from the map and a
  style sentence (insets), generated frame at 3.8\,s. (b)~Held-out test worlds; insets show the input map.}
  \label{fig:app-worlds}
\end{figure*}

\begin{figure*}[t]
  \centering
  \includegraphics[width=\linewidth]{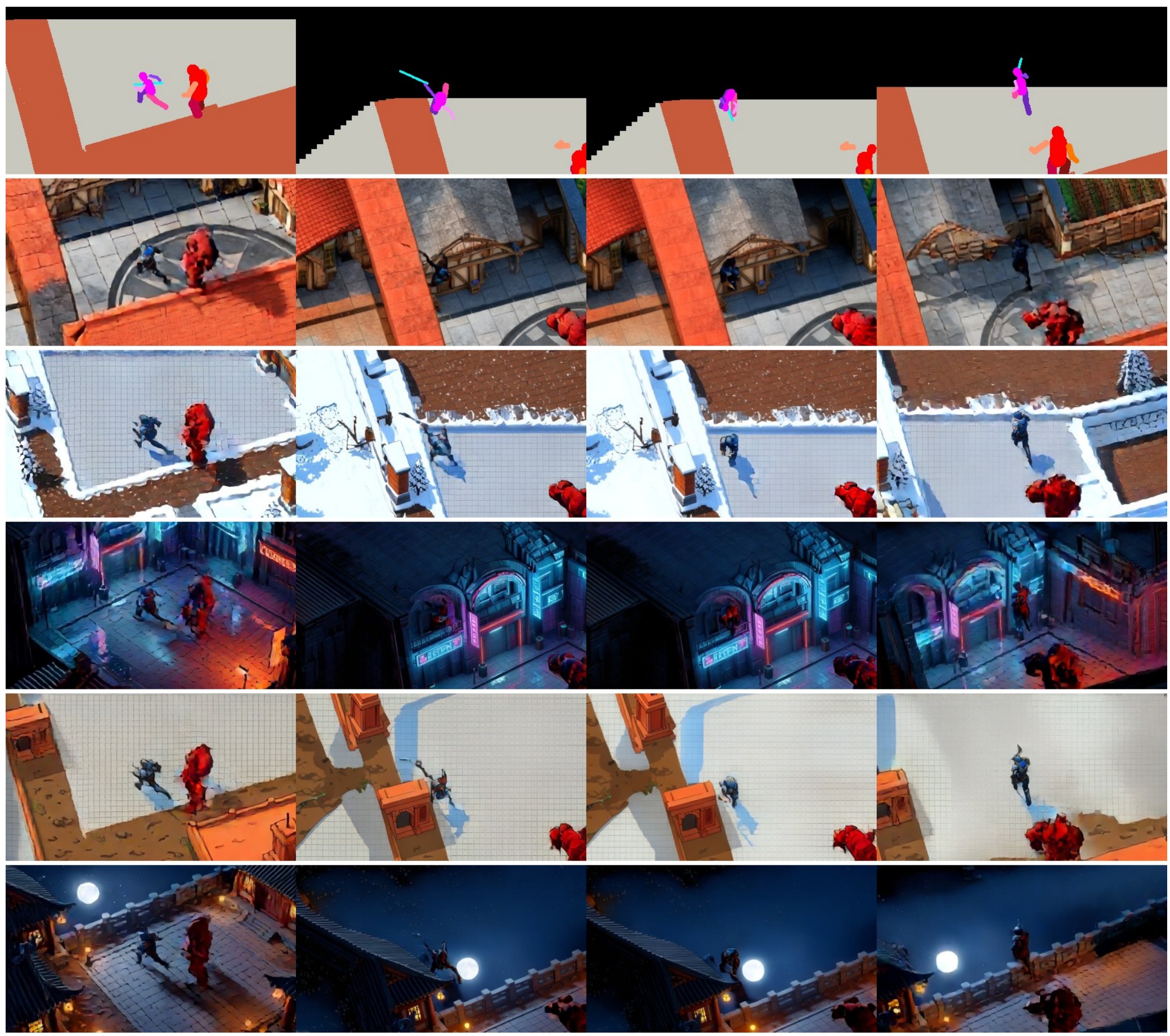}
  \caption{\textbf{Same map, five first frames.} Top: the map of one recorded demo session at 1, 4, 7 and 10\,s (black:
  never observed). Rows: the student replaying exactly this map sequence from the recorded first frame (medieval) and
  from four restyled first frames (winter, cyberpunk, cartoon, night).}
  \label{fig:app-samemap}
\end{figure*}

Fig.~\ref{fig:app-worlds}(a) extends Fig.~\ref{fig:teaser}(c) to nine styles. One procedurally generated map that does
not occur in training is started from nine first frames that FLUX.2 drew from the map and a one-sentence style; the
student then generates all 81 frames from the map sequence. Large images show the frame at 3.8\,s and insets the first
frame. Fig.~\ref{fig:app-worlds}(b) shows held-out test worlds with their recorded maps. Fig.~\ref{fig:app-samemap}
repeats the test on a recorded play session of the browser demo (Sec.~\ref{sec:app-demo}): the exact map sequence of
that session is replayed through the student five times, from its recorded first frame and from four first frames to
which the demo's image-edit model (Qwen-Image-Edit) restyled the recorded first frame. Maps are identical in every step;
layout, characters and timing follow the map, materials and lighting the first frame.

\section{Unseen maps vs.\ SolarRenderer}
\label{sec:app-unseen}
\begin{table}[t]
\centering
\caption{\textbf{Unseen worlds}: mIoU (\%) on procedurally generated maps in four themes (10 each) and on 20 maps
painted by a programmatic freehand sketch generator, rendered with test-clip cameras and characters and a first frame
generated from the map. In-dist.: the test set of Tab.~\ref{tab:main}.}
\label{tab:app-unseen}
\small
\setlength{\tabcolsep}{3pt}
\resizebox{\columnwidth}{!}{%
\begin{tabular}{lcccccc}
\toprule
Method & In-dist. & Canyon & Forest & Plaza & Village & Sketch \\
\midrule
SolarRenderer & 62.7 & 44.1 & 50.6 & 32.5 & 49.1 & 37.2 \\
(1) colour + VAE & 65.4 & 54.7 & 60.3 & 39.8 & 48.9 & 37.9 \\
(2) colour + light & 66.7 & 55.7 & 63.5 & 39.4 & 50.7 & 38.7 \\
(3) integer, teacher & 65.8 & 55.5 & 63.0 & 39.7 & 45.1 & 33.5 \\
\textbf{MaLiang student} & 62.3 & \textbf{56.1} & \textbf{64.9} & 39.1 & 49.5 & 38.7 \\
\bottomrule
\end{tabular}}
\end{table}

\begin{figure*}[t]
  \centering
  \includegraphics[width=\linewidth]{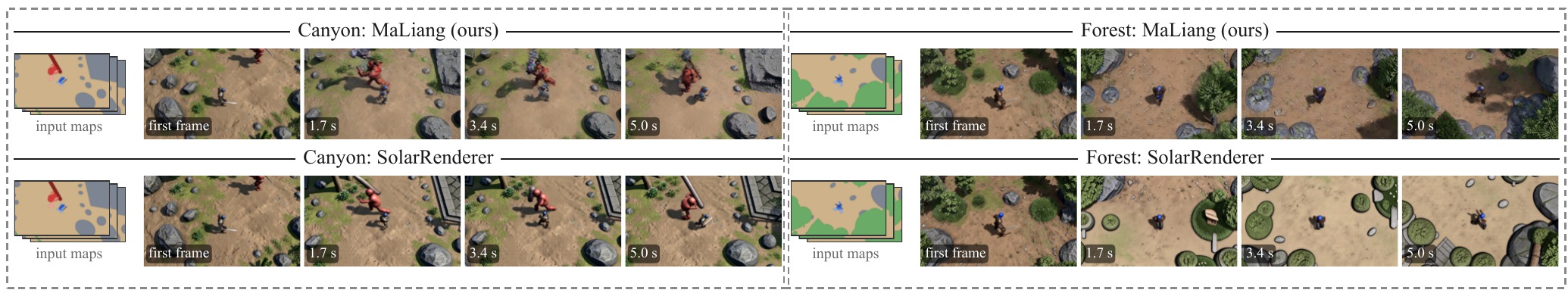}
  \caption{\textbf{Unseen worlds}, MaLiang student vs.\ SolarRenderer from the same first frame.}
  \label{fig:app-unseen}
\end{figure*}

Fig.~\ref{fig:app-unseen} shows two of the procedurally generated worlds scored in Tab.~\ref{tab:app-unseen}. Both
methods start from the same first frame. MaLiang keeps the canyon and forest layouts over the clip, whereas SolarRenderer
replaces the forest floor with a different scene within two seconds.

\section{Interactive demo}
\label{sec:app-demo}
\begin{figure*}[t]
  \centering
  \includegraphics[width=\linewidth]{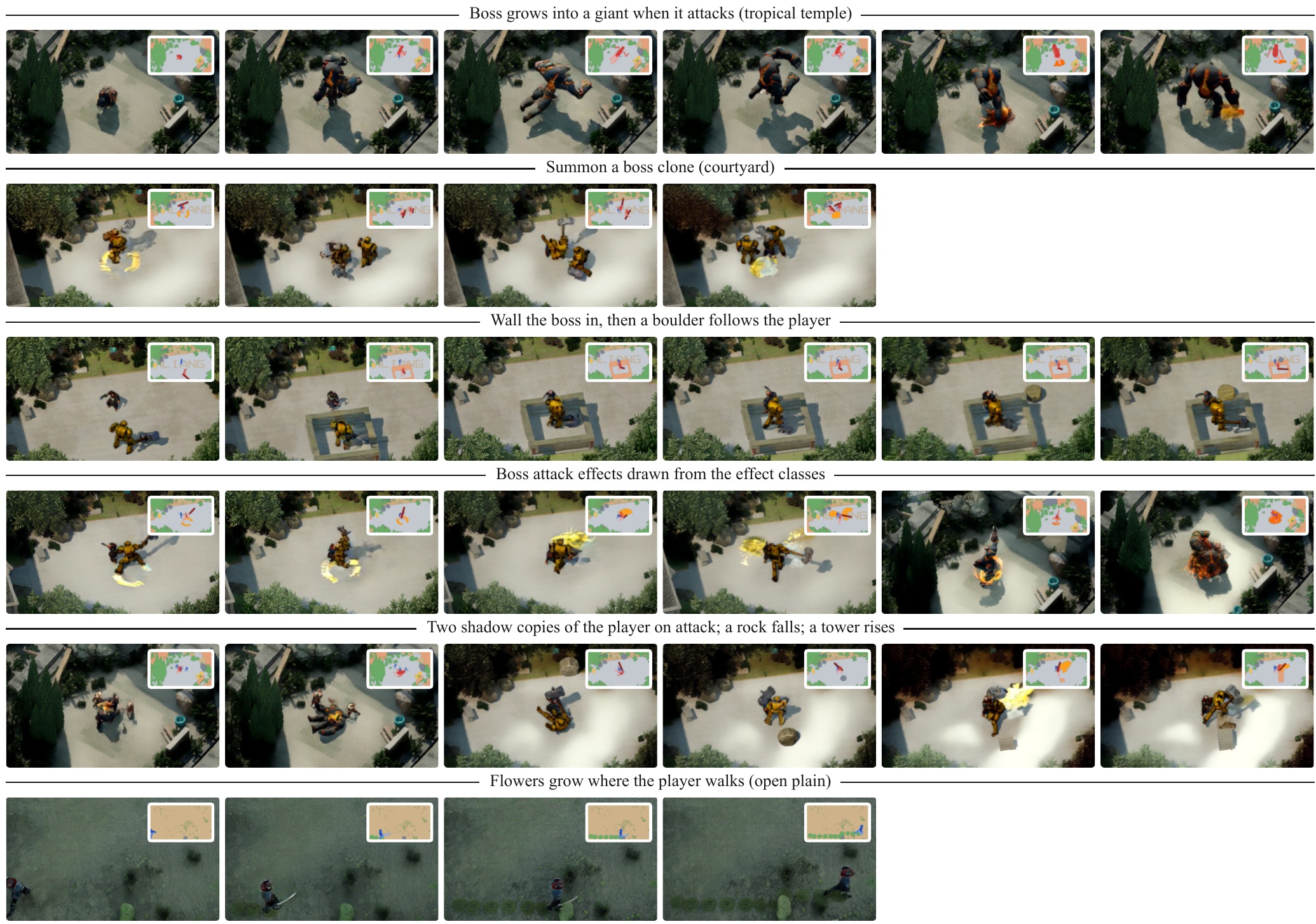}
  \caption{\textbf{Interactive demo sequences} generated by the deploy student while the game edits the map.}
  \label{fig:app-demo}
\end{figure*}

Our browser demo runs a small game whose state is rasterised into the integer map every 0.25\,s and streamed to the
student; it adds game rules that write into the map: a boss clone, walls around the boss, a boulder that follows the
player, a falling rock, a rising tower, flowers along the player's path, a boss that grows when it attacks, shadow copies
of the player, and the attack-effect classes. Fig.~\ref{fig:app-demo} shows longer sequences (frames in temporal order,
a few seconds per row; insets: the map the student received).

\section{Live editing and effect control}
\label{sec:app-edit}
Tab.~\ref{tab:app-edit} gives the full numbers behind Sec.~\ref{sec:edit}, split by edit type and including the
offline teachers.

\paragraph{All 100 test clips.} Tab.~\ref{tab:app-editall} repeats the edit test on every test clip with the same edit
construction, edit time and scores, now per target class and object size. An added object is an upright cylinder
2.6\,m high written into the map at 2.3\,s; a removal replaces one static category by the ground class it covers. Rocks
are drawn in about two thirds of their pixels at every size and appear within a few frames; removals are rendered as the
uncovered ground in 92\% (terrain) and 53\% (paving) of their pixels. Plants and buildings respond less; a solid upright
cylinder resembles the compact rocks of the recordings far more than their vegetation, which is fragmented. Outside the
edit the video stays as in the unedited run (33--36\,dB for added objects).

\begin{table}[t]
\centering
\caption{\textbf{Live editing} (33 world-anchored edits on held-out clips, applied at 2.3\,s; averaged over edits) and
\textbf{effect control} (94 held-out clips with boss attacks, paired runs with and without effect classes; deploy
student). ``Without'': same pixels, unedited run.}
\label{tab:app-edit}
\small
\setlength{\tabcolsep}{4pt}
\resizebox{\columnwidth}{!}{%
\begin{tabular}{lccc}
\toprule
\emph{Edits}: pixels rendered as the new class & without & with & gain \\
\midrule
Add an object (rock / plant / building), $n{=}21$ & 1.5\% & 57.9\% & $+56.4$ \\
Remove a category, $n{=}12$ & 30.8\% & 85.0\% & $+54.2$ \\
All 33, live student & 13.6\% & 69.1\% & $+55.5$ \\
All 33, teacher (1) / (2) / (3), offline & 11.6 / 11.5 / 11.2 & 67.9 / 69.9 / 70.3 & \\
\midrule
Unedited region vs.\ no edit, PSNR / LPIPS & \multicolumn{3}{c}{31.5\,dB / 0.055} \\
Visible in the generated video after & \multicolumn{3}{c}{1.3 frames} \\
\midrule
\emph{Effects} & \multicolumn{3}{c}{} \\
\midrule
Pixels changed inside the effect region & \multicolumn{3}{c}{63.3\%} \\
Pixels changed outside (49\,px margin) & \multicolumn{3}{c}{1.3\%} \\
Frames where effect is drawn iff in the map & \multicolumn{3}{c}{93.8\%} \\
\bottomrule
\end{tabular}}
\end{table}

\begin{table}[t]
\centering
\caption{\textbf{Live editing on all 100 test clips} (deploy student). Pixels inside the edit rendered as the target
class, mean over the last 2\,s; ``without'': the unedited run on the same pixels. $n$: clips where the edit is visible in
the map at the end (at least 50 pixels). Visible: frames after the edit until half of the final value is reached. PSNR:
unedited region (49\,px margin) against the unedited run.}
\label{tab:app-editall}
\small
\setlength{\tabcolsep}{4pt}
\resizebox{\columnwidth}{!}{%
\begin{tabular}{lrrrrr}
\toprule
Edit & $n$ & without & with & visible & PSNR \\
\midrule
Add rock, radius 0.8\,m & 67 & 1.4\% & \textbf{64.1\%} & 0.3 & 36.1 \\
Add rock, radius 1.2\,m & 70 & 1.4\% & \textbf{63.5\%} & 3.9 & 35.2 \\
Add rock, radius 2.0\,m & 75 & 2.8\% & \textbf{52.9\%} & 3.4 & 33.2 \\
Add vegetation, radius 1.2\,m & 70 & 5.4\% & 31.0\% & 5.8 & 35.4 \\
Add building, radius 1.2\,m & 70 & 5.2\% & 9.4\% & 12.4 & 35.7 \\
\midrule
Remove, uncovers terrain & 58 & 38.7\% & \textbf{91.9\%} & 0.0 & 28.7 \\
Remove, uncovers paving & 23 & 4.3\% & \textbf{53.0\%} & 3.5 & 32.4 \\
\bottomrule
\end{tabular}}
\end{table}

\section{Condition cost}
\label{sec:app-cost}
Tab.~\ref{tab:app-cost} breaks down the per-step cost summarised in Sec.~\ref{sec:costexp}.

\begin{table}[t]
\centering
\caption{\textbf{Condition cost} on one GPU at 832$\times$480, deploy student, per four-frame step (0.25\,s of video).
Median (95th percentile) over repeated runs. m: measured, e: estimated by replacing only the condition path (the
transformer is the same).}
\label{tab:app-cost}
\small
\setlength{\tabcolsep}{3pt}
\resizebox{\columnwidth}{!}{%
\begin{tabular}{lccc}
\toprule
 & Integer map & \multicolumn{2}{c}{Colour proxy +} \\
\cmidrule(lr){3-4}
 & (ours) & video VAE & TAE \\
\midrule
Condition data / frame, PNG & 4.9\,KB & \multicolumn{2}{c}{10.5\,KB} \\
\quad as compressed array (used) & 2.4\,KB & \multicolumn{2}{c}{--} \\
Condition encoder params & 9.4M & 150M & 1.5M + 9.4M$^{\ast}$ \\
Condition encode, ms (m) & 0.91 (0.98) & 49.4 & 1.9 \\
Transformer step incl.\ cond., ms & 180 (188) m & 228 e & 181 e \\
Decode, TAEHV, ms (m) & 3.5 (3.6) & 3.5 & 3.5 \\
Throughput incl.\ decode, fps & 21.8 & 17.2 e & 21.7 e \\
\bottomrule
\end{tabular}}\\[2pt]
{\footnotesize $^{\ast}$A latent proxy still needs a projection into the model. Throughput assumes the transformer step
and decoding run one after the other ($4/0.1836\,$s $=21.8$\,fps); overlapping them would give 22.2\,fps.}
\end{table}

\section{Three sampling seeds}
\label{sec:app-seeds}
The interface ablation (Tab.~\ref{tab:ablation}) samples every teacher with seed 0. We repeated the full protocol with seeds
1 and 2: the same 100 test clips, first frames, captions and 50-step sampler, and the 94-clip effect test.
Tab.~\ref{tab:app-seeds} reports the mean and spread over the three seeds and, for every metric, in how many seeds the
difference is significant under the significance rule of Sec.~\ref{sec:setup} (clip and recording bootstrap both exclude 0, tested per seed). The two
advantages of integer conditioning hold in every seed: position IoU is higher than with either colour interface (+0.011
to +0.020), and the effect-region response and timing are higher than with the VAE path (+7.7 to +9.5 and +5.1 to +6.5
points). Against the light colour encoder (2) the effect gains are significant in two of three seeds (+1.8 to +3.5
points; in seed 1 only the clip-level interval excludes 0). Over the VAE path the temporal-stability gain also holds in
all seeds, the time-match gain only in seed 0; (2) keeps its small SSIM / LPIPS edge in all seeds. Layout (mIoU), FVD
and hallucination rate stay tied, as in Tab.~\ref{tab:ablation}.

\begin{table*}[t]
\centering
\caption{\textbf{Interface ablation over three seeds} (teachers, Tab.~\ref{tab:ablation} protocol; seed 0 is the run in Tab.~\ref{tab:ablation}). Mean
$\pm$ standard deviation over seeds 0, 1, 2. Bottom: number of seeds in which (3) is significantly better (\up) or worse
(\dn) than the colour interface; 0: tied in every seed.}
\label{tab:app-seeds}
\setlength{\tabcolsep}{3pt}
\resizebox{\textwidth}{!}{%
\begin{tabular}{lccccccccccc}
\toprule
 & PSNR$\uparrow$ & SSIM$\uparrow$ & LPIPS$\downarrow$ & FVD$\downarrow$ & mIoU$\uparrow$ & Time$\uparrow$ & Pos.$\uparrow$
 & Halluc.$\downarrow$ & Warp$\downarrow$ & Resp.$\uparrow$ & Timing$\uparrow$ \\
\midrule
(1) colour $\to$ video VAE & 20.45$\pm$0.08 & 0.503$\pm$0.001 & 0.305$\pm$0.002 & 112$\pm$2 & 65.8$\pm$0.4
 & 0.866$\pm$0.003 & 0.563$\pm$0.005 & 12.3$\pm$0.7 & 9.88$\pm$0.21 & 52.9$\pm$1.5 & 88.5$\pm$1.4 \\
(2) colour $\to$ light enc. & 20.59$\pm$0.03 & 0.520$\pm$0.003 & 0.294$\pm$0.001 & 115$\pm$1 & 66.4$\pm$0.3
 & 0.871$\pm$0.003 & 0.567$\pm$0.004 & 12.1$\pm$0.4 & 8.80$\pm$0.24 & 58.9$\pm$1.6 & 92.4$\pm$1.5 \\
\textbf{(3) integer (ours)} & 20.53$\pm$0.14 & 0.513$\pm$0.004 & 0.299$\pm$0.002 & 116$\pm$2 & 65.7$\pm$0.3
 & \textbf{0.878}$\pm$0.003 & \textbf{0.580}$\pm$0.002 & 11.8$\pm$0.3 & 8.80$\pm$0.27 & \textbf{61.4}$\pm$1.3
 & \textbf{94.2}$\pm$0.7 \\
\midrule
(3) vs.\ (1), seeds significant & 2\up & 2\up & 1\up & 0 & 0 & 1\up & \textbf{3\up} & 0 & \textbf{3\up} & \textbf{3\up}
 & \textbf{3\up} \\
(3) vs.\ (2), seeds significant & 0 & 3\dn & 3\dn & 0 & 0 & 0 & \textbf{3\up} & 0 & 0 & 2\up & 2\up \\
\bottomrule
\end{tabular}}
\end{table*}

\section{Entity size}
\label{sec:app-size}
Sec.~\ref{sec:ablation} explains the position advantage by pooling: a colour code averages a thin limb with its
surroundings, whereas class fractions keep its share. Tab.~\ref{tab:app-size} tests this on the renders of Tab.~\ref{tab:ablation} by
splitting every frame's player and boss by their area in the map. Integer conditioning gives the highest position IoU in
every size bin, and its lead is largest for small characters (500--1500\,px: +0.040 over (1), +0.041 over (2)). It is
significant over (1) from 500\,px up and over (2) from 1500\,px up; the smallest bin (288 frames) is too small to decide.

\begin{table}[t]
\centering
\caption{\textbf{Characters by size} (player and boss, area of their map mask in pixels at 832$\times$480; teachers of
Tab.~\ref{tab:ablation}, seed 0). Position IoU of the generated entity with its map mask, and detection (the segmenter finds the entity on
its mask). $^{\ast}$: (3) significantly higher than both colour interfaces (clip and recording bootstrap);
$^{\dagger}$: than (1) only.}
\label{tab:app-size}
\small
\setlength{\tabcolsep}{3pt}
\resizebox{\columnwidth}{!}{%
\begin{tabular}{lrccccccc}
\toprule
 & & \multicolumn{3}{c}{Position IoU} & & \multicolumn{3}{c}{Detection, \%} \\
\cmidrule(lr){3-5}\cmidrule(lr){7-9}
Area, px & frames & (1) & (2) & (3) & & (1) & (2) & (3) \\
\midrule
150--500 & 288 & 0.298 & 0.315 & \textbf{0.335} & & 13.9 & 15.6 & 14.6 \\
500--1500 & 583 & 0.408 & 0.407 & \textbf{0.448}$^{\dagger}$ & & 55.4 & 60.5 & 59.7 \\
1500--4000 & 4285 & 0.556 & 0.562 & \textbf{0.579}$^{\ast}$ & & 84.6 & 85.9 & 86.6$^{\dagger}$ \\
$>$4000 & 5438 & 0.576 & 0.581 & \textbf{0.595}$^{\ast}$ & & 91.5 & 92.2 & 92.4 \\
\bottomrule
\end{tabular}}
\end{table}

\section{Imperfect maps}
\label{sec:app-imperfect}
A game may deliver an imperfect map. Tab.~\ref{tab:app-imperfect} runs the deploy student on the 100 test clips with
perturbed input maps and scores the output against the recording, as in Tab.~\ref{tab:main}; two more seeds on the clean
map give the run-to-run spread. Scattered wrong pixels (up to 5\%) and a map at a quarter of the resolution stay within
that spread. A map shifted by 4\,px (0.5\% of the width) costs 0.017 position IoU, by 8\,px 0.058. Unknown regions (the
\emph{unknown} class, which the model sees in training for parts of the world the camera has not mapped yet) lower the
scores with the hidden share. A delayed map costs the most, because the video follows the map: the output for a map 4
frames late matches the clean-map output of 4 frames earlier at 24.6\,dB but the same frame at only 19.9\,dB, in all 100
clips (8 frames: 24.4 vs.\ 18.8\,dB). The drop therefore measures the lag itself, and a game should send the map of the
current step.

\begin{table}[t]
\centering
\caption{\textbf{Imperfect maps} (deploy student, 100 test clips, scored against the recording). Noise: each map pixel
replaced by a random class of the clip with the given probability. Shift: whole map translated in a random direction.
Low-res.: map rasterised at 208$\times$120 and upsampled. Unknown: random rectangles covering the given share of the
frame set to \emph{unknown} for the whole clip. Delay: map lags the game by 4 frames (one streaming step, 0.25\,s) or 8.}
\label{tab:app-imperfect}
\small
\setlength{\tabcolsep}{3pt}
\resizebox{\columnwidth}{!}{%
\begin{tabular}{lccccccc}
\toprule
Input map & PSNR$\uparrow$ & LPIPS$\downarrow$ & FVD$\downarrow$ & mIoU$\uparrow$ & Time$\uparrow$ & Pos.$\uparrow$
 & Halluc.$\downarrow$ \\
\midrule
Clean & 20.63 & 0.327 & 153 & 62.2 & 0.892 & 0.586 & 12.1 \\
\quad seed 1 & 20.76 & 0.326 & 157 & 61.3 & 0.889 & 0.584 & 12.5 \\
\quad seed 2 & 20.70 & 0.326 & 148 & 61.4 & 0.892 & 0.582 & 13.2 \\
\midrule
Noise 1\% & 20.58 & 0.327 & 150 & 62.2 & 0.893 & 0.585 & 12.4 \\
Noise 5\% & 20.52 & 0.334 & 149 & 61.7 & 0.886 & 0.583 & 12.6 \\
Low-res.\ 1/4 & 20.42 & 0.331 & 153 & 62.1 & 0.891 & 0.581 & 12.2 \\
Shift 4\,px & 20.10 & 0.333 & 150 & 61.1 & 0.889 & 0.569 & 11.6 \\
Shift 8\,px & 19.47 & 0.347 & 152 & 58.4 & 0.883 & 0.528 & 11.5 \\
Unknown 10\% & 20.17 & 0.342 & 201 & 56.2 & 0.838 & 0.551 & 15.0 \\
Unknown 30\% & 19.76 & 0.360 & 239 & 50.9 & 0.789 & 0.500 & 15.6 \\
Delay 4 frames & 17.80 & 0.454 & 164 & 43.0 & 0.789 & 0.418 & 15.7 \\
Delay 8 frames & 17.24 & 0.480 & 179 & 37.3 & 0.737 & 0.407 & 17.5 \\
\bottomrule
\end{tabular}}
\end{table}

\section{More comparisons on held-out clips}
\label{sec:app-qual}
\begin{figure*}[t]
  \centering
  \includegraphics[width=\linewidth]{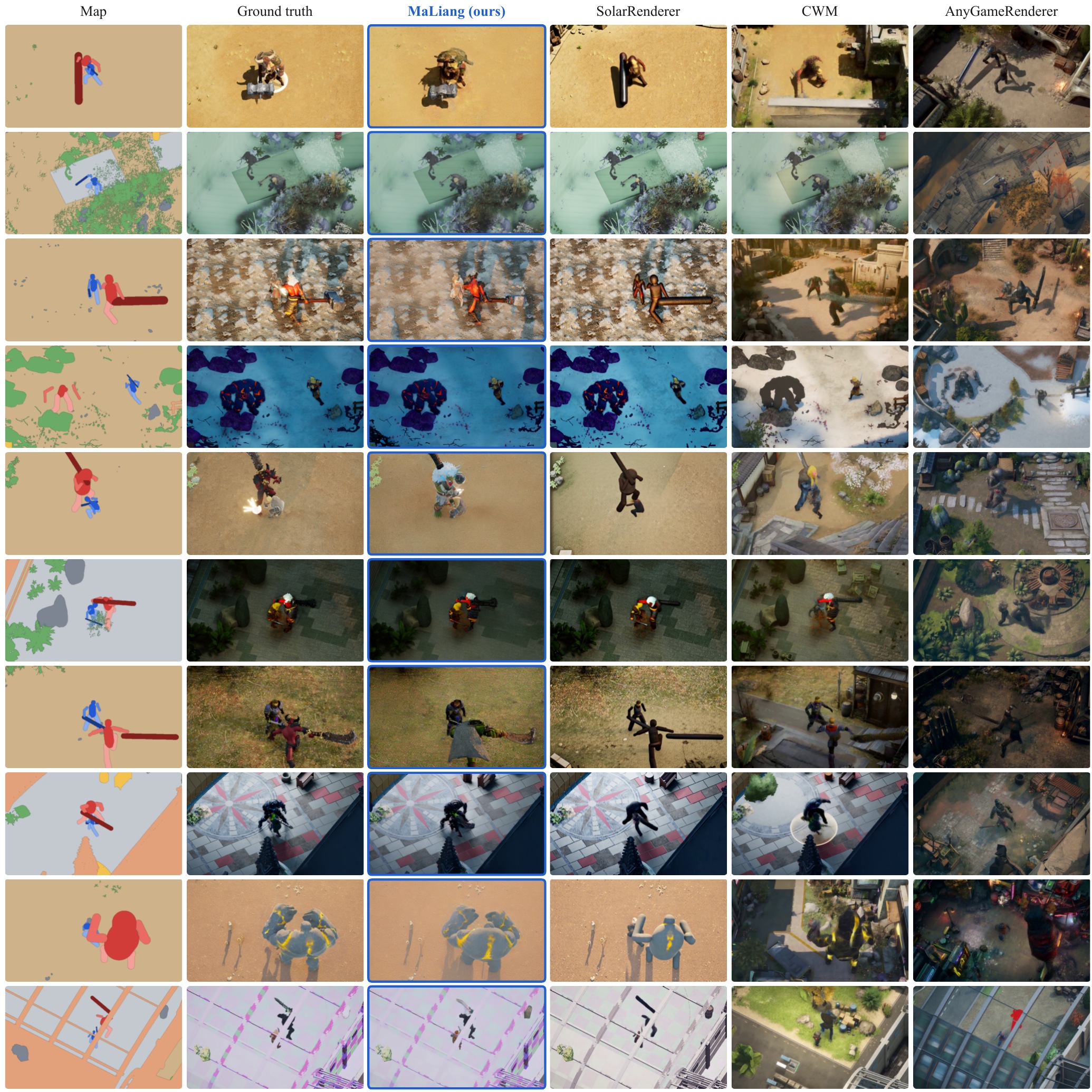}
  \caption{\textbf{One held-out clip per test scene} (largest median boss area), frame at 2.5\,s. Baselines are run off
  the shelf, MaLiang is the deploy student.}
  \label{fig:app-qual}
\end{figure*}

Fig.~\ref{fig:app-qual} extends Fig.~\ref{fig:qual} to all ten test scenes. For each scene we take the clip with the
largest median boss area (a fixed rule, not a selection of good results) and show frame 40 (2.5\,s) of every method.

\section{More same-backbone examples}
\label{sec:app-abl}
\begin{figure*}[t]
  \centering
  \includegraphics[width=\linewidth]{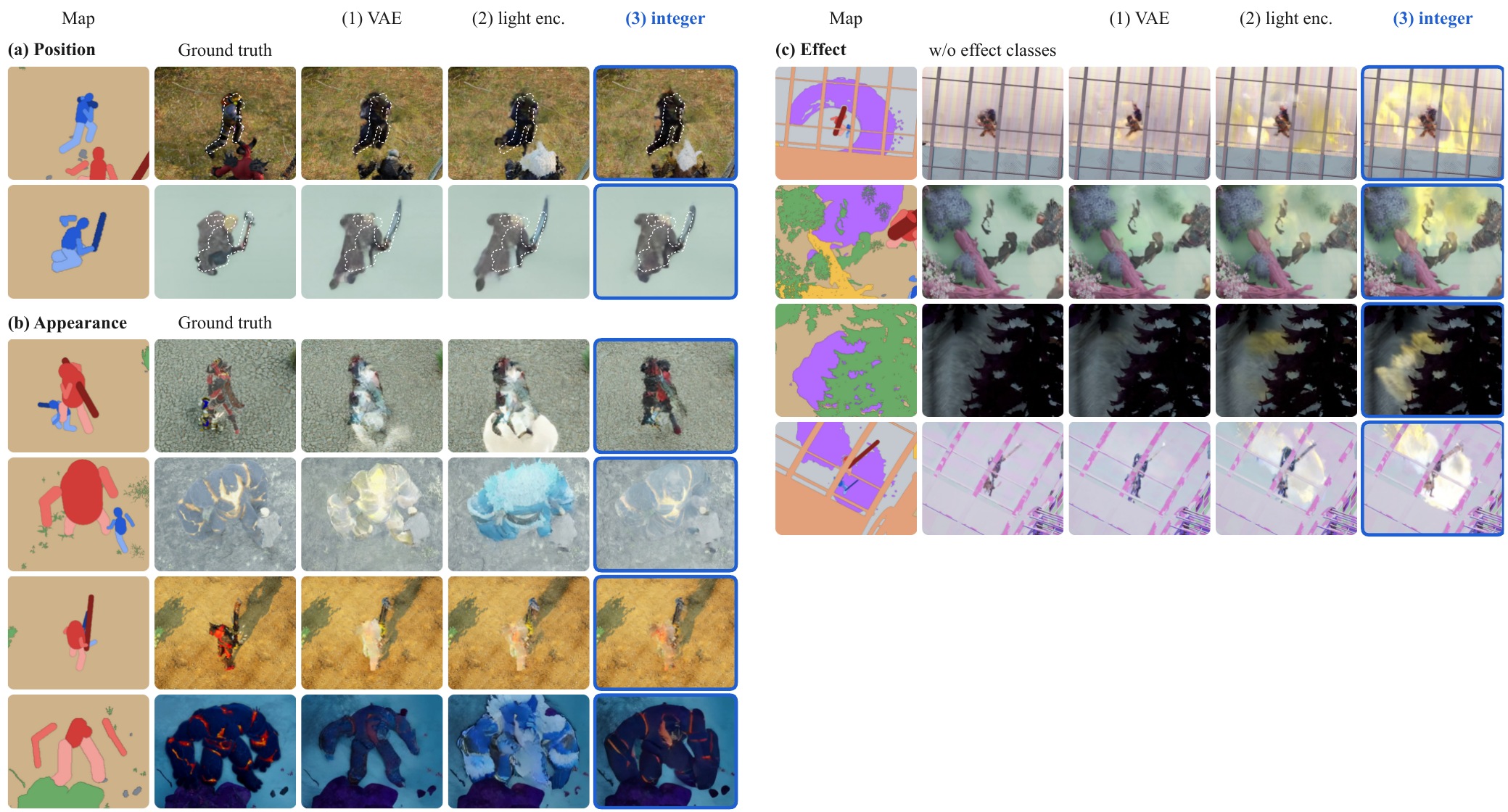}
  \caption{\textbf{Same backbone, different interface: more examples.} (a) Position: dashed white outline = the player's
  mask in the map. (b) Appearance: the crystal golem, the armoured knights and the lava golem stay closest to the
  recording with integer conditioning. (c) Boss-attack effects added to the maps of recordings that contain none, so
  there is no ground truth; the second column is the integer teacher on the original map.}
  \label{fig:app-abl}
\end{figure*}

Fig.~\ref{fig:app-abl} adds examples to Fig.~\ref{fig:ablcrops} for the three teachers of the interface ablation (same
backbone, data and training steps; same seed and first frame).

\section{One-minute rollouts}
\label{sec:app-long}
\begin{figure*}[t]
  \centering
  \includegraphics[width=0.6\linewidth]{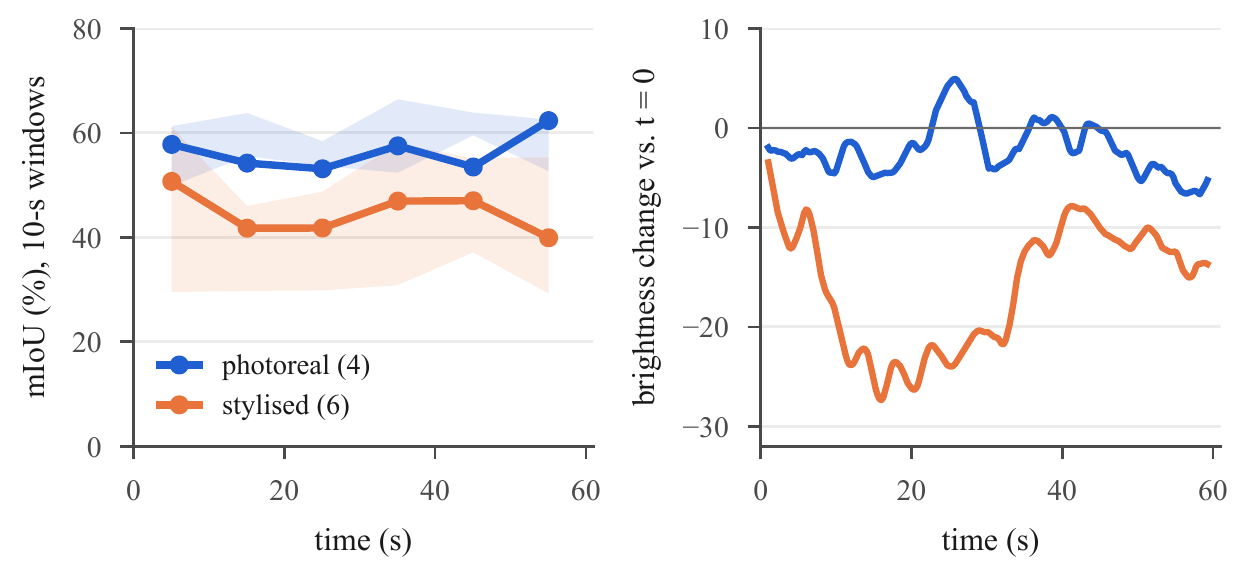}
  \caption{\textbf{One-minute rollouts} (10 chains of 961 frames, held-out recordings). Left: pooled mIoU per 10-s
  window (band: inter-quartile range over chains). Right: brightness relative to frame 0.}
  \label{fig:app-long}
\end{figure*}

Fig.~\ref{fig:app-long} shows the curves behind the long-horizon numbers of Sec.~\ref{sec:general}: ten chains of 961
frames (241 streaming steps) on consecutive held-out clips, with the recorded maps and captions.

\end{document}